\documentclass[10pt,journal,compsoc]{IEEEtran}
\ifCLASSOPTIONcompsoc
  \usepackage[nocompress]{cite}
\else
  \usepackage{cite}
\fi
\usepackage{url}
\usepackage[utf8]{inputenc} % allow utf-8 input
\usepackage[T1]{fontenc}    % use 8-bit T1 fonts
\usepackage{booktabs}       % professional-quality tables
\usepackage{amsfonts}       % blackboard math
\usepackage{amsthm}
\usepackage{nicefrac}       % compact symbols for 1/2, etc.
\usepackage{microtype}      % microtypography
\usepackage{xcolor}         % colors
\usepackage{amsmath}
\usepackage{subfigure}
\usepackage{graphicx}
\usepackage{enumitem}
\usepackage{mathtools}
\usepackage{wrapfig}
\usepackage{algorithm}
\usepackage{algorithmic}
\usepackage{multirow}
\newtheorem{remark}{Remark}

\newtheorem{proposition}{Proposition}
\newtheorem{theorem}{Theorem}

\usepackage{comment}
\usepackage{bm}
\usepackage{makecell}

\usepackage{tikz}
\newcommand*\circled[1]{\tikz[baseline=(char.base)]{
    \node[shape=circle,draw,inner sep=0.5pt] (char) {#1};}}

\usepackage[utf8]{inputenc}

\def\cpar{\hss\egroup\line\bgroup\hss}

\font\tenrm = cmr16 at 10pt

\def\ro{\textcolor{black}}
\def\rt{\textcolor{black}}

\ifCLASSINFOpdf
\else
\fi
\begin{document}
\title{Earning Extra Performance from Restrictive Feedbacks\\
%More Efficacy from Restrictive Feedbacks
%Smaller Update is More Efficacious for Restrictive Feedbacks\\
%Smaller Update on Deep Models from Restrictive Feedbacks
%Layerwise Update without Regret from Restrictive Feedbacks
}
\author{Jing Li, %~\IEEEmembership{Student Member}
        Yuangang Pan,
        Yueming Lyu,
        Yinghua Yao,
        Yulei Sui,
        and~Ivor W. Tsang,~\IEEEmembership{Fellow,~IEEE}
\IEEEcompsocitemizethanks{
\IEEEcompsocthanksitem Jing~Li is with the Australian Artificial Intelligence Institute, University of Technology Sydney, Australia, and also with the Center for Frontier AI Research, Agency for Science, Technology and Research (A*STAR), Singapore. \protect\\
E-mail: j.lee9383@gmail.com
\IEEEcompsocthanksitem Yuangang~Pan is with the Center for Frontier AI Research, Agency for Science, Technology and Research (A*STAR), Singapore. \protect\\
Email: yuangang.pan@gmail.com
\IEEEcompsocthanksitem Yueming~Lyu is with the Center for Frontier AI Research, Agency for Science, Technology and Research (A*STAR), Singapore.\protect\\
Email: yueming.lyu@gmail.com
\IEEEcompsocthanksitem Yinghua~Yao is with the Guangdong Key Laboratory of Brain-inspired Intelligent Computation, Department of Computer Science and Engineering, Southern University of Science and Technology, Shenzhen, China,~518055 and also with the Australian Artificial Intelligence Institute, University of Technology Sydney, Australia,~2007.\protect\\
E-mail: yinghua.yao@student.uts.edu.au
\IEEEcompsocthanksitem Yulei~Sui is with School of Computer Science and Engineering, University of New South Wales, Sydney, Australia.\protect\\
Email: y.sui@unsw.edu.au
\IEEEcompsocthanksitem Ivor~W.~Tsang is with the Center for Frontier AI Research, Agency for Science, Technology and Research (A*STAR), Singapore.\protect\\
Email: ivor.tsang@gmail.com.}
%\thanks{Main part of this work was done while the first author was at the Southern University of Science and Technology.}
\thanks{Accepted by IEEE TPAMI in April 2023.}
}


\markboth{Journal of \LaTeX\ Class Files,~Vol.~xx, No.~x, August~2022}%
{Shell \MakeLowercase{\textit{et al.}}: Bare Demo of IEEEtran.cls for Computer Society Journals}



\IEEEtitleabstractindextext{%
\begin{abstract}
Many machine learning applications encounter a situation where model providers are required to further refine the previously trained model so as to gratify the specific need of local users. This problem is reduced to the standard model tuning paradigm if the target data is permissibly fed to the model. However, it is rather difficult in a wide range of practical cases where target data is not shared with model providers but commonly some evaluations about the model are accessible. In this paper, we formally set up a challenge named \emph{Earning eXtra PerformancE from restriCTive feEDdbacks} (EXPECTED) to describe this form of model tuning problems. Concretely, EXPECTED admits a model provider to access the operational performance of the candidate model multiple times via feedback from a local user (or a group of users). The goal of the model provider is to eventually deliver a satisfactory model to the local user(s) by utilizing the feedbacks. Unlike existing model tuning methods where the target data is always ready for calculating model gradients, the model providers in EXPECTED only see some feedbacks which could be as simple as scalars, such as inference accuracy or usage rate.
To enable tuning in this restrictive circumstance, we propose to characterize the geometry of the model performance with regard to model parameters through exploring the parameters' distribution. In particular, for the deep models whose parameters distribute across multiple layers, a more query-efficient algorithm is further tailor-designed that conducts layerwise tuning with more attention to those layers which pay off better. Our theoretical analyses justify the proposed algorithms from the aspects of both efficacy and efficiency. Extensive experiments on different applications demonstrate that our work forges a sound solution to the EXPECTED problem, which establishes the foundation for future studies towards this direction. Code is available via \textcolor{red}{https://github.com/kylejingli/EXPECTED}.%, such as inference accuracy or usage rate.
\end{abstract}

%for crafting more promising descendants.

}

\maketitle

\IEEEdisplaynontitleabstractindextext

% \fi
\IEEEpeerreviewmaketitle

\IEEEraisesectionheading{\section{Introduction}\label{sec:introduction}}


% 1.Previous GAN conditions on labels (TeachGAN; Latent constraint)
% 1-1 GAN an unsupervised generative model
\IEEEPARstart{P}{lainly} using a provided model usually cannot fulfill the requirement of downstream tasks, probably on account of the distribution shift on data~\cite{popov2018distributed,wang2020fully}, or altered evaluation metrics~\cite{adel2019one}. For example, after the deployment of a language model on user devices, model updates are often needed to enable a stronger performance on personalized data~\cite{popov2018distributed}. These necessary model updates, known as model tuning in the machine learning community, compensate for the potential discrepancy between upstream and downstream tasks. With accessible target data to the deployed model, the end-to-end back-propagation based model tuning has been demonstrated as a powerful technique in a wide range of fields, such as computer vision~\cite{donahue2014decaf,chen2020big}, natural language~\cite{howard2018universal, DBLP:conf/naacl/DevlinCLT19}, and medicals~\cite{tajbakhsh2016convolutional}. However, there remain many model tuning applications that cannot be dealt with in this manner. %We present a practical example here to show this form of cases.
\begin{figure*}[!t]
    \centering
    \includegraphics[width=0.9\textwidth]{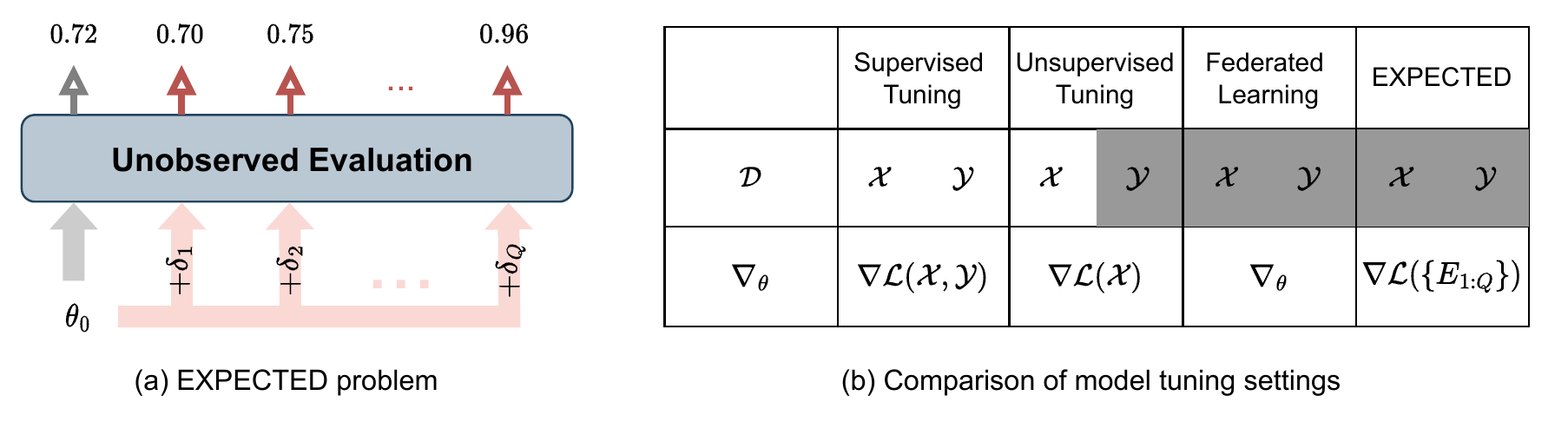}\vskip-0.2in
    \caption{\label{fig:Intro} Overview of EXPECTED. (a) Given a deployed model parameterized by $\theta_0$, EXPECTED aims to adapt it to the target task with limited query-feedbacks (budget $Q$) through the unobserved evaluation. (b)~We instance the unobserved evaluation by the inaccessibility of target data. In this case, EXPECTED is compared with other three model tuning settings from the aspects of (1)~how much information about target data $\mathcal{D}$ is accessible and (2)~how the gradient information $\nabla_{\theta}$ is attained. The grey filling indicates the object is unobserved to the tuning executor. In term of the federated learning, although local data $\mathcal{X},\mathcal{Y}$ is inaccessible to the global model, the gradient $\nabla_\theta$ is directly returned. Note that $E_i$ is informally short for $E(\mathcal{D};(\theta_0+\delta_i))$.}%For example, can we perturb $\theta_0$ within $Q$ times to achieve the desired test performance?
\end{figure*}

\emph{{Example.}} (Model tuning with hided target data) Alice possesses a model developed on a collection of data (public or private), and she would only send out an application instead of the source model (e.g., to protect intellectual property). Bob is a user who owes local data which might be different from Alice's. He is also unwilling to share his data (e.g., personal concerns) but wishes Alice could update the source model so that the corresponding application eventually achieves a pleasant performance for him. This requirement is rational because the users who experiences unsatisfied performance will become discouraged and more likely to quit using the application~\cite{hashimoto2018fairness}. Typically, Bob returns some feedback about how the candidate model performs, assisting Alice in doing meaningful product updates. Such interactions are normally included in mutually agreed protocols in the real world~\cite{rashidi2009keeping}. \rt{Please refer to more real scenarios such as 
tuning personalized foundation models~\cite{brown2020language} and learning without third-party 
platform~\cite{zhai2016cqstr} in Appendix.} %\rt{We present two scenarios including tuning personalized foundation models~\cite{brown2020language} and learning without third-party 
%

%Data privacy researchers have offered some workarounds to this dilemma. For example, one can send out encrypted data~\cite{DBLP:journals/iotj/ChamikaraBKLCA20} to achieve a trade-off between data utility and privacy. However, this approach is impractical for the non-expert data holder, and also the informative feedbacks are overlooked. 
Beyond privacy motivation, there are other cases for which the standard model tuning is not applicable either. For example, in a user-centric system, the perceptual evaluation that reflects personal and situational characteristics is invaluable to boost the system~\cite{knijnenburg2012explaining}. These subjective evaluations (e.g., user rating) show the desire for model efficacy and cannot be naturally treated as input data for the tuning purpose. Despite the specific scenarios, we summarize the common trait of all the involved applications. That is, although the model provider cannot access target data for standard model tuning, the model is still hopefully to be improved by merely utilizing some downstream feedback information.

In this study, we abstract this form of learning problems as \emph{Earning eXtra PerformancE from restriCTive feEDdbacks}, dubbed by EXPECTED. Each feedback in EXPECTED is an evaluation result of a legal candidate model, thus our tuning objective is to achieve a satisfactory evaluation result on the target task. In this sense, ``earning extra performance'' means the improvement of evaluation performance over the initially provided model after multiple queries. ``Restrictive'' has twofold meanings here. On one hand, the model evaluation result should be uncomplicated, because the common evaluation metric like inference accuracy is typically a score and the user subjective evaluation is probably a star-rating. On the other hand, the number of evaluations is supposed to be limited due to the practical requirement for communication cost or efficiency. Fig.~\ref{fig:Intro}(a) depicts the EXPECTED problem, where the given model $\theta_0$ is to be tuned so as to achieve a performance as high as possible with $Q$ queries. To make it more understandable, we use ``unobserved evaluation'' to absorb different cases of the aforementioned applications.

EXPECTED challenges the existing research and poses a new and difficult model tuning task. Without the explicit target data which has been treated as an indispensable ingredient for learning, model tuning cannot be executed through the standard back-propagation implemented in different software repositories. In addition, although one can design some heuristic strategies to guess what a better model for the target task looks like from feedbacks, conducting a gradient-based optimization for model tuning remains troublesome, especially for modern Deep Neural Networks (DNNs) which are in high dimensional space and often with complex structures.

Intuitively, the past evaluated models provide valuable experience for crafting the latter ones. For instance, if a model is said poorly behaved from the feedback, then the current query model should explore in a different direction. This insight suggests solving the EXPECTED problem not only involves cherishing the valuation of feedback information but also focuses on designing approaches to instruct the generation of candidate models. Having this principle in mind, we propose to refine the provided model by characterizing the geometry of evaluation performance with regard to model parameters. 
Based on natural evolution strategy (NES)~\cite{wierstra2014natural}, we develop the efficient gradient-based model tuning algorithms for both lightweight models and complex DNNs. Our theoretical analyses and experiments justify the utility of the proposed algorithms. %show that the deployed model is progressively optimized . %Specifically, the newly crafted model is searched along the direction in which the most historical models succeed improving.

The main contributions of this work are summarized as follows.
\begin{itemize}
\item Motivated by many real applications which demands further updates to the previously trained model from restrictive model evaluations, we introduce the setting of Earning eXtra PerformancE from restriCTive feEDdbacks (EXPECTED). EXPECTED is not a conventional data-driven optimization problem and thus supplements the existing model tuning regime. We also clarify the difference between EXPECTED and other model tuning paradigms to highlight its novelty.
\item We realize that EXPECTED problem could be effectively remedied if historical model evaluations are elaborately designed to provide valuable clues. Based on this understanding, we propose Performance-guided Parameter Search~(PPS) algorithm which resorts to optimizing the distribution of model parameters via gradient estimation. In terms of tuning DNNs, Layerwise Coordinate Parameter Search~(LCPS) algorithm is further brought forward to significantly reduce the query number. Plus, we theoretically justify the soundness of our algorithms.
\item The experiments on different modality data, including tabular data, text, and images, demonstrate the efficacy of the proposed algorithms for both data distribution shifts and altered evaluation metrics. We also verify that LCPS adaptively prioritizes more useful layers to update which saves the query cost. In particular, we find that our method that only uses restrictive feedbacks even rivals the unsupervised tuning works~\cite{sun2020test,wang2020fully} that access the entire features of target data on corrupted image classification task, showing a great potential in applications where features are not shared out.
\end{itemize}

\section{EXPECTED Compared with Other Learning Settings}
Adapting a pre-trained model to the related target tasks motivates the model tuning setting. In this work, we focus on discrimintive models. For a better statement, we assume the inaccessible target data only differs source data by (input) distribution shift (We leave the case of altered metrics in Section \ref{sec:beyond}), and thus the network architecture does not need modifying. Fig.~\ref{fig:Intro}~(b) compares EXPECTED with the three most related model tuning settings from the following two aspects. 

\textbf{1) Model tuning setting evolves with more restrictive target information accessible.} If sufficient target data including features and labels are accessible during tuning, this is literally the supervised learning paradigm, i.e., fine-tuning~\cite{donahue2014decaf}. Once the label information is absent, it comes to the unsupervised tuning, which is also known as the test-time training~\cite{sun2020test,wang2020fully} or source-free unsupervised domain adaptation~\cite{sahoo2020unsupervised,li2020model,liang2020we}. Federated learning~\cite{mcmahan2017communication} lets the decentralized global model fit local data without sharing them out. While studied in a one-to-many context, it can be viewed as a model tuning process from global to local\footnote{The involved global-local interaction strictly becomes a model tuning process when a collaborator has data changes and aims to acquire a fine-tuned model from the global model, referred by the recent work~\cite{mazumder2021restricted}.}. However, one can see that federated learning preserves local data by bringing model training to the device, which is in fact not applicable to the scenarios where intellectual property is also concerned as referred in the previous example.
Uniquely, the proposed EXPECTED neither accesses features nor labels of target data, and it only allows limited two-way communications, i.e., querying with model candidates and receiving model performances as feedbacks. 
% Recent works~\cite{li2021improved,shachaf2021theoretical} are completing fine-tuning by proposing insightful theoretical instructions.  ... The slight difference between them is that test-time training treats test data as target in an online stream while source-free UDA aims to transfer the source model to fit the whole target distribution under a source data free circumstance.

\textbf{2) More restrictive target information implies a harder gradient-based optimization.} Through the above statement, we notice that both supervised tuning and federated learning actually take the sufficient gradient information because they are empirically derived on the labelled target data. In terms of unsupervised tuning where only $\mathcal{X}$ is observed, model gradient is computed from the self-supervision formulation~\cite{sun2020test} or in a fashion of self-training~\cite{yarowsky1995unsupervised}. Therefore, their gradients might be biased (See CIFAR-10-C experimental results in Section~\ref{sec:shifted_data_distribution}). We can see all these three settings can be conclusively categorized to the data-driven model tuning, as their gradients can be sample-wise decomposed. Things for EXPECTED are indirect by contrast, because the ingredients of EXPECTED for computing gradients are query models and their feedbacks. By understanding every feedback $E_i$ as a summary statistic of the target data in terms of the $i$-th query model, EXPECTED is consequently interpreted as a model-driven tuning problem. In particular, we expect that our proposed algorithms achieve the compared performance with the data-driven model tuning methods, even when the query budget is not very generous.

\section{Preliminaries}
We first prepare the common mathematical notations used in this paper as Table.~\ref{table:notations}.

\textbf{Problem Formulation.} Let $F_{\theta_0}$ denote the initially provided (pre-trained) model which is parameterized with ${\theta}_0$, $\mathcal{D}$ denote the inaccessible target data that the model aims to adapt to, and $Q$ denote query budget, i.e., the tolerant number of model evaluations. For a probing candidate model $F_{\theta_i} (1\le i \le Q)$, evaluation function $E(\cdot)$ measures its performance over target data $\mathcal{D}$ and returns a score $s_i$ (When multiple evaluation metrics are used, a tuple might be returned of which $s_i$ could be as an element. One can refer to Section~\ref{sec:fair_classification} for a case study about this scenario.) as feedback. That is, $s_i=E(\mathcal{D};F_{\theta_i})$. Supposing a larger score is preferred, e.g., accuracy, EXPECTED aims to solve the following problem,
\begin{equation}\label{eq:original}
    \theta_*= \arg\max_\theta E(\mathcal{D};F_\theta), \;\; s.t.\; {\#queries} \le Q.
\end{equation}
Please also see Fig.~\ref{fig:Intro}(a) for this example. Note that we will use an alternative form $E(\mathcal{D};\theta)$ or $E(\theta)$ to replace $E(\mathcal{D};F_\theta)$ for the convenient expression in the rest of the paper when it does not cause any ambiguity.

\textbf{A Naive Approach -- Random Search.}
With query chance budgeted by $Q$, one can randomly perturb the deployed model and ask for its evaluation on the target data. Afterwards, the model that achieves the best performance is selected as the optimal approximation of $\theta_*$. That is, 
\begin{equation}\label{eq:theta_plus}
    \theta_* \approx \theta_0 + \arg\max_{\delta_i}\{E(\mathcal{D};\theta_0+\delta_i)\}_{i=1}^Q,
\end{equation}
where $\delta_i$ represents the difference between the initially provided model $F_{\theta_0}$ and the tuned model $F_{\theta_i}$. If each $\delta_i$ is derived independently, solving Eq.~\eqref{eq:theta_plus} comes to a Random Search~\cite{bergstra2012random} game, which may not meet the need of aforementioned restrictive conditions as the parameter space
is too large to do a search in this way (See Section~\ref{sec:shifted_data_distribution}). 

\noindent\emph{Notice}. Transferring other emerging hyperparameter searching techniques~\cite{buczak2021using} or advanced evolution algorithms~\cite{opara2019differential} to the model tuning scenario is beyond our scope. This study will focus on how to effectively solve the EXPECTED problem based on gradient estimation, especially when modern DNNs are to be tuned. 

\begin{table}[t]
\caption{\label{table:notations} Common mathematical notations.} %\vskip-0.07in
\centering
\begin{tabular}{l|l}
\toprule
 Notation&Explanation  \\ \midrule
 $\mathcal{D}=\{\mathcal{X},\mathcal{Y}\}$& target data with features $\mathcal{X}$ and labels $\mathcal{Y}$\\
  $Q$& query budget  \\
 $H$ & layer number of tuned parameters\\
 $\theta_i$ & model parameter for query index $i$\\
 $\theta^t$ & model parameter updated after $t$-th iteration\\
 $F_\theta(\cdot)$ &  model $F$ parameterized with $\theta$ \\ 
 $E(\cdot)$ & evaluation metric\\
 $G(\cdot)$ & performance gain by tuning\\
 $\pi(\cdot)$&  distribution of model parameters  \\
 $l_h \in \theta$&  parameters of the $h$-th layer  \\
 $p_h$&  probability of the $h$-th layer to be sampled \\
 $u$ & the number of queries for a unit update\\
 $b$& batch size of samplings  \\
 $\sigma$& standard variance  \\
 $\epsilon \sim \mathcal{N}(0,I)$& standard Gaussian noise  \\
 $|\cdot|$ & dimension of a vector\\
 $||\cdot||$& $\ell_2$-norm of a vector\\
 \bottomrule
\end{tabular}
\end{table}

\section{Tuning from Restrictive Feedbacks}
This section starts with a general case in which we aim to optimize $\theta$ by taking advantage of the feedback information. With the consideration of model structures, the second part focuses on tuning DNNs under EXPECTED. \rt{We leave the computation cost analyses of our methods to Appendix.}
\subsection{Gradient-based Optimization from Query-feedbacks}
The constraint about the query number in objective~\eqref{eq:original} can be simply eliminated by applying a stopping criterion about the performance gain. For convenience, we treat it as an unconstrained problem by still running the full $Q$ queries. 

%Our intuition for solving EXPECTED problem is that humans craft new models from the experience on old ones. Keeping this thought in mind, our goal is to formulate EXPECTED as a gradient-based optimization problem.
\subsubsection{Learning the Distribution of Model Parameters}
If the evaluation function $E(\cdot)$ of EXPECTED denotes the classification accuracy, its specific form is then written as
\begin{equation}\label{eq:0_1_loss}
 E(\mathcal{D};\theta) = 1-
 \frac{1}{|\mathcal{D}|} \sum_{(x_i,y_i) \in \mathcal{D}} \mathbb{I}(F_{\theta}(x_i) \neq y_i),
\end{equation}
where $\mathbb{I}(\cdot)$ is the sign function, a.k.a. zero-one loss. The right hand term is the negative expression of empirical risk, which suggests that problem~\eqref{eq:original} can be decomposed over the target samples, i.e., 
\begin{equation}\label{eq:tuning_loss}
    \max_{\theta}\mathbb{E}_{x,y\sim p_{(x,y)}}E(x,y;\theta).
\end{equation}
Problem~\eqref{eq:tuning_loss} can be viewed as the standard tuning paradigm with the indifferentiable loss. As $p(x,y)$ is agnostic, we alternatively consider the following form via introducing the distribution of $\theta$,
\begin{equation}
    E(\mathcal{D};\theta) =  E(\mathcal{D};\mathbb{E}_{\theta \sim \pi(\theta)}(\theta)) \approx \mathbb{E}_{\theta \sim \pi(\theta)}E(\mathcal{D};\theta),
\end{equation}
where $\theta$ is assumed sampled from the parameter distribution $\pi(\theta)$, the equality holds by defining $\theta$ as its expectation over $\pi(\theta)$, and the later approximation follows~\cite{wierstra2014natural}. Since we intend to obtain the optimal $\theta_*$, solving problem~\eqref{eq:original} can be written as
\begin{equation}\label{eq:samplings}
\begin{aligned}
    \theta_*  \sim &\pi_*(\theta),\\
    \pi_*(\theta) =  \arg\max_{\pi(\theta)} & \mathbb{E}_{\theta \sim \pi(\theta)} E(\mathcal{D};\theta),
\end{aligned}
\end{equation}
where $\pi_*(\theta)$ represents the best estimation of $\pi(\theta)$, and the expectation is taken over all candidate models. We can see that this proxy objective relaxes the optimization to $\theta$ into characterizing its distribution $\pi(\theta)$. 

To make it practical, we parameterize $\pi$ by the density probability $\pi(\theta|\omega)$, where $\omega$ denotes the distribution parameters. As a result, solving problem~\eqref{eq:samplings} requires the maximization of the following objective,
\begin{equation}\label{eq:expectation_fit}
    J(\omega)=\mathbb{E}_{\theta \sim \pi(\theta|\omega)}[E(\mathcal{D};\theta)]=\int E(\mathcal{D};\theta) \pi(\theta|\omega)\,d\theta.
\end{equation}
The gradient of Eq.~(\ref{eq:expectation_fit}) w.r.t. $\omega$ can be computed by 
\begin{equation}\label{eq:gradient}
\begin{aligned}
\nabla_\omega J(\omega) & \overset{\circled{1}}{=} \mathbb{E}_{\theta \sim \pi(\theta|\omega)}[E(\mathcal{D};\theta)\nabla_\omega \log \pi(\theta|\omega)] \\
&\overset{\circled{2}}{\approx} \frac{1}{b}\sum_{i=1}^b E(\mathcal{D};\theta_i)\nabla_\omega \log \pi(\theta_i|\omega),
\end{aligned}
\end{equation} 
where $\circled{1}$ uses the so-called log-likelihood trick~\cite{wierstra2014natural} which enables the gradient decoupled from the evaluation function $E(\cdot)$, $\circled{2}$ adopts the Monte Carlo approximation by empirically conducting $b$ samplings from $\pi(\theta|\omega)$, i.e., $\theta_1,..., \theta_b$. From Eq.~\eqref{eq:gradient}, we can see that the involved samplings $\theta_i$ for estimating the gradient of $\omega$ can be properly achieved by query chances of EXPECTED. Concretely, in every iteration, we consume $b$ queries to draw from the current $\pi(\theta|\omega)$ which are used to estimate the gradient of $\omega$. Then we update $\omega$ by the gradient ascent and obtain a new $\pi(\theta|\omega)$ for the next round of samplings. This process will not terminate until $Q$ queries run out or some candidate is already satisfactory.

\subsubsection{Implementation with Gaussian Prior}
Recall that the canonical form of training a supervised model is i.i.d. log-likelihoods plus a log prior. It is known that the widely used weight regularizer from literature is the weight decay, which corresponds to a centered Gaussian prior. Following this convention, we instance the distribution of the model parameter $\pi(\theta|\omega)$, i.e., $\omega=[\mu, \Sigma]$. Optimizing $\omega$ requires the natural gradient for scale conformity~\cite{wierstra2014natural} which generally involves the inverse of the {Fisher information matrix} with the size of $|\theta| \times |\theta|$ in our case. To reduce the burden of heavy computation, we assume that  $\Sigma \approx \sigma^2I$ will not hurt the tuning performance by much, which enables us to treat $\sigma^2$ as a hyperparameter and only to estimate the gradient of $\theta$.
For example, at the first iteration, $\mu$ is initialized by $\theta^0$ ($\theta^0:=\theta_0$), i.e., $\pi(\theta|\omega) = \mathcal{N}{(\theta|\theta^0,\sigma^2I)}$. By leveraging the reparameterization technique, the candidate models are sampled around the current model $\theta^0$, i.e., $\theta_i=\theta^0+\sigma\epsilon_i \; (1\le i\le b)$, where $\epsilon_i \sim \mathcal{N}(0,I)$. Taking such samplings into Eq.~\eqref{eq:gradient} yields the gradient estimation w.r.t. $\omega$. As updating $\omega$ is equivalently updating $\theta$, \rt{we then directly provide the gradient estimation with sampling batch size $b$ w.r.t. $\theta$,}
\begin{equation}\label{eq:for_sgd_update}
 \nabla \mathbb{E} [E(\theta)] \approx \frac{1}{\sigma b}\sum_{i=1}^b \epsilon_iE(\theta+\sigma\epsilon_i),
\end{equation}
where $\mathcal{D}$ is dropped from now on for simplicity. Although starting from a surrogate objective~\eqref{eq:samplings}, the above implementation helps us optimize model parameters $\theta$ directly. 

Before taking Eq.\eqref{eq:for_sgd_update} into a gradient ascent update, we exhibit two techniques to facilitate this gradient estimation. First, we adopt antithetic sampling~\cite{geweke1988antithetic} which is demonstrated to stabilize the update. That is, in each round, we only independently sample $b/2$ Gaussian points $\epsilon_j$ and let the rest be the negative copies, i.e., $\epsilon_{b-j+1} = -\epsilon_j$. Second, to reduce the impact of the scale of model performance, we normalize the feedbacks by subtracting the mean and then dividing by the standard deviation before using them. Such a normalization step has been demonstrated to maintain a constant learning rate $\eta$~\cite{li2019nattack} and also provides an important condition for some fundamental facts used in Appendix. Alg.~\ref{alg1} summarizes the whole procedure, which is named Performance-guided Parameters Search (PPS).

\begin{algorithm}[tb]
        \caption{Performance-guided Parameter Search (PPS)}\label{alg1}
        \begin{algorithmic}[1]
          \REQUIRE Initially provided model $F_{\theta_0}$, query budget~$Q$, learning rate $\eta$, batch size~$b$, variance~$\sigma^2$.
          \FOR {$t=0,..., \lfloor Q/b\rfloor$}
          \STATE Sample $\{\epsilon_j\}_{j=1}^{b/2} \sim \mathcal{N}(0,I)$, and for each $j$ get $\epsilon_{b-j+1} = -\epsilon_j$.
          \STATE Generate candidate models $\{\theta_i\}_i^b$ as queries where $\theta_i = \theta^t+\sigma\epsilon_i$. \textcolor{gray}{ $ \# \theta^t=\theta_0$ if $t=0$}
          \STATE Collect and normalize feedbacks $\{E(\mathcal{D};\theta_i)\}_{i=1}^b$.
          \STATE $\theta^{t+1} \gets \theta^{t}+\frac{\eta}{\sigma b}\sum_{i=1}^b \epsilon_iE(\mathcal{D};\theta^t+\sigma\epsilon_i)$.
          \ENDFOR
          \ENSURE $\theta^{\lfloor Q/b\rfloor+1}$.
        \end{algorithmic}
\end{algorithm}

\subsubsection{Quality Analysis of the Estimated Gradient}
%We theoretically analyze the effectiveness of the estimated gradient by quantifying its . 
We first show that applying the antithetic sampling on Eq.~\eqref{eq:for_sgd_update} allows us to explicitly build the connections between the estimated gradient and the true gradient.
\begin{proposition}\label{prop:1}
If $\sigma$ is small, any estimated gradient $\nabla\mathbb{E}[E(\theta)]$ derived by Alg.~\ref{alg1} can be seen as a projection of the corresponding true gradient $\nabla E(\theta) \in \mathbb{R}^{|\theta|}$ onto a lower dimensional space with $b/2$ independent random Gaussian vectors being bases. 
\end{proposition}
\begin{proof}
When antithetic sampling is used, the expression of Eq.~\eqref{eq:for_sgd_update} can be written as:
\begin{equation*}
\begin{aligned}
\nabla\mathbb{E}[E(\theta)] &\approx \frac{1}{\sigma b} \sum_{i=1}^b E(\theta+\sigma \epsilon_i)\epsilon_i\\
&\overset{\circled{1}}{=}\frac{1}{b/2}\sum_{i=1}^{b/2} \frac{ E(\theta+\sigma \epsilon_i)-E(\theta-\sigma \epsilon_i)}{2\sigma}\epsilon_i\\
&\overset{\circled{2}}{\approx} \frac{1}{b/2}\sum_{i=1}^{b/2} \left(\nabla_{\epsilon_i} E(\theta) \cdot ||\epsilon_i|| \right) \epsilon_i\\
&\overset{\circled{3}}{=}\frac{1}{b/2}\sum_{i=1}^{b/2} \langle \nabla E(\theta), \epsilon_i \rangle \epsilon_i\\
&\overset{\circled{4}}{=}\frac{1}{b/2}\sum_{i=1}^{b/2} \text{Proj}_{\epsilon_i} (\nabla E(\theta)) \cdot||\epsilon_i|| \cdot  {\epsilon_i},
\end{aligned}
\end{equation*}
where $\circled{1}$ follows the step 2 of Alg~\ref{alg1}, $\circled{2}$ uses the definition of directional derivative when $\sigma \rightarrow 0$, $\circled{3}$ rewrites the directional derivation into a form of the dot product, and $\circled{4}$ is a natural reformulation to align with the definition of vector projection. Taking all the independent $\epsilon_i$ as bases, the coordinate value of $\nabla\mathbb{E}[E(\theta)]$ onto each base is $\text{Proj}_{\epsilon_i} (\nabla E(\theta)) \cdot||\epsilon_i||$, which completes the proof.
\end{proof}

We then introduce the following theorem which quantifies how well a random projection preserves the length information of a vector.
\begin{theorem}\label{theorem1}
Let $M \in \mathbb{R}^{ |\theta| \times \frac{b}{2}}$ denote the random projection matrix with $||\epsilon_i||\cdot\epsilon_i \quad (i=1,2,...,\frac{b}{2})$ being the columns. For the true gradient $\nabla E(\theta)$ at any $\theta$, we have
\begin{equation*}\small
\begin{aligned}
    &\Pr \left\{ (1-\xi)||\nabla E(\theta)||^2 \le ||M^T\nabla E(\theta)||^2 \le (1+\xi)||\nabla E(\theta)||^2 \right\}\\ 
    & \qquad\qquad >1-2e^{-C\xi^2b} ,
\end{aligned}
\end{equation*}
where $\xi \in (0,1)$ and $C >0$ is a constant. \hfill\rule{2mm}{2mm}
\end{theorem}
Theorem~\ref{theorem1} is a direct application of the Johnson-Lindenstrauss Lemma~\cite{matouvsek2013lecture} but uses the unnormalized Gaussian bases. The norm of projected coordinates is lower and upper bounded, which means the length of true gradient $\nabla E(\theta)$ is almost preserved after the projection $M$. Let $a \in \mathbb{R}^{\frac{b}{2}}$ denote the coefficient vector with the $i$-th entry being $\text{Proj}_{\epsilon_i} (\nabla E(\theta)) \cdot||\epsilon_i||$. As any $\epsilon_i, \epsilon_j$ are nearly orthogonal~\cite{gorban2016approximation}, according to Proposition~\ref{prop:1} we have
\begin{equation}\label{eq:threom1}
    ||\nabla\mathbb{E}[E(\theta)]||^2 \approx ||a||^2. 
\end{equation}
Since $a=M^T\nabla E(\theta)$, from Theorem~\ref{theorem1} we conclude that our estimated gradient almost preserves the length of corresponding true gradients.
\begin{remark}
The approximation in Eq.~\eqref{eq:threom1} hinders the strict comparison between $||\nabla E(\theta)||^2$ and $||\nabla\mathbb{E}[E(\theta)]||^2$. However, generally speaking, when $b$ increases, this approximation is more accurate as the fidelity is better preserved, and the bound of Theorem~\ref{theorem1} becomes tighter as well.
\end{remark}
 
\subsubsection{Toy Example for PPS}\label{sec:toy_example}
We present a toy example to verify the efficacy of Alg.~\ref{alg1}. Three-layer perceptron networks ($3$-MLP) are firstly pre-trained on source data (two Gaussians with the variance of $[0.7,0.7]$) and are then tuned following Alg.~\ref{alg1} on target data -- another two Gaussians with the variance of $[0.1,1.5]$). In this experiment, we only tune the last layer of the $3$-MLP. The evaluation error on half of randomly selected target data is used as the feedback, and the query budget is set as $80$.

\begin{figure}[ht]
\centering
    \subfigure[Pre-training on source]{\label{fig:a}\includegraphics[width=0.23\textwidth]{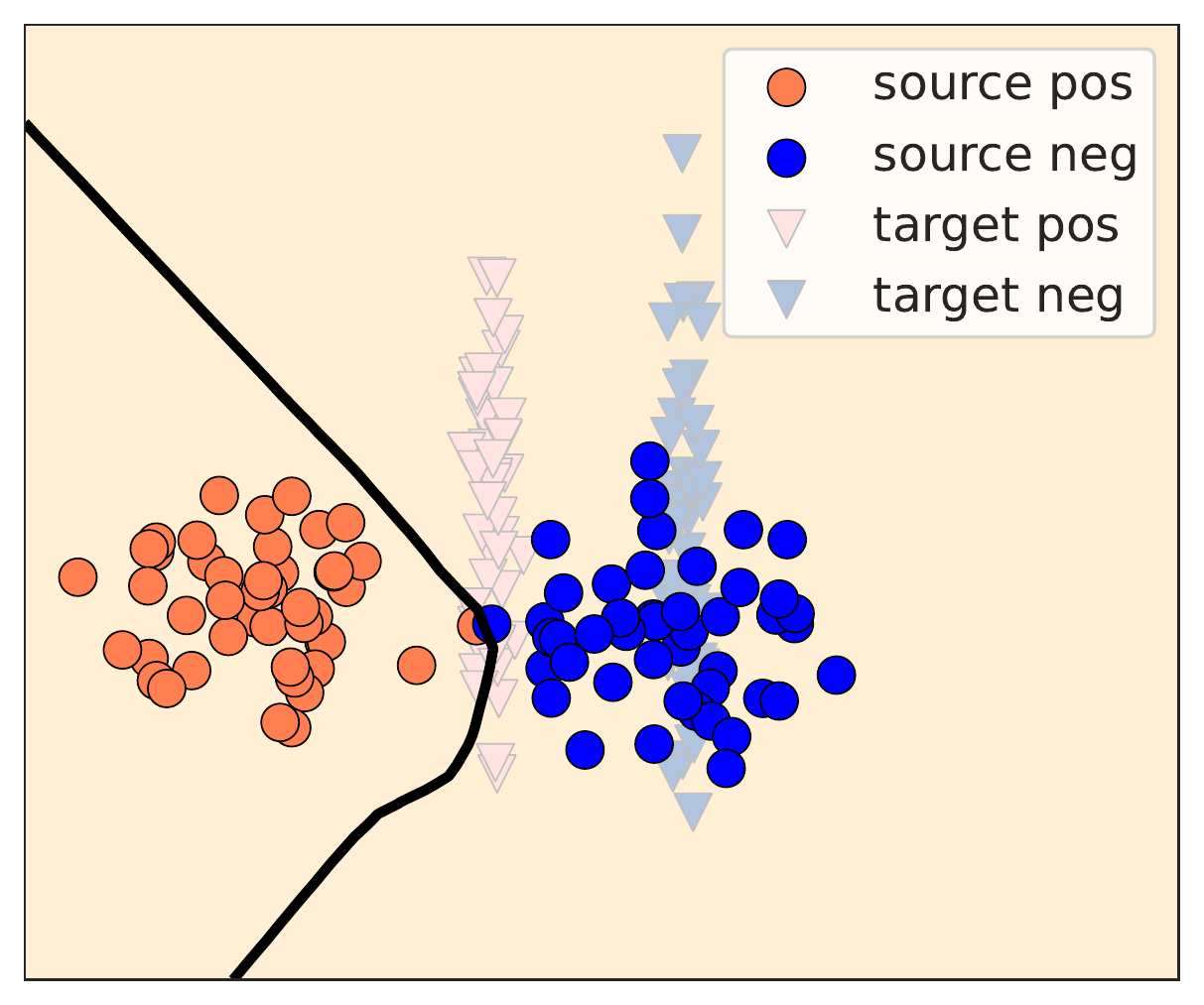}}
    \hfill
    \subfigure[PPS for target ($Q=80$) ]{\label{fig:b}\includegraphics[width=0.23\textwidth]{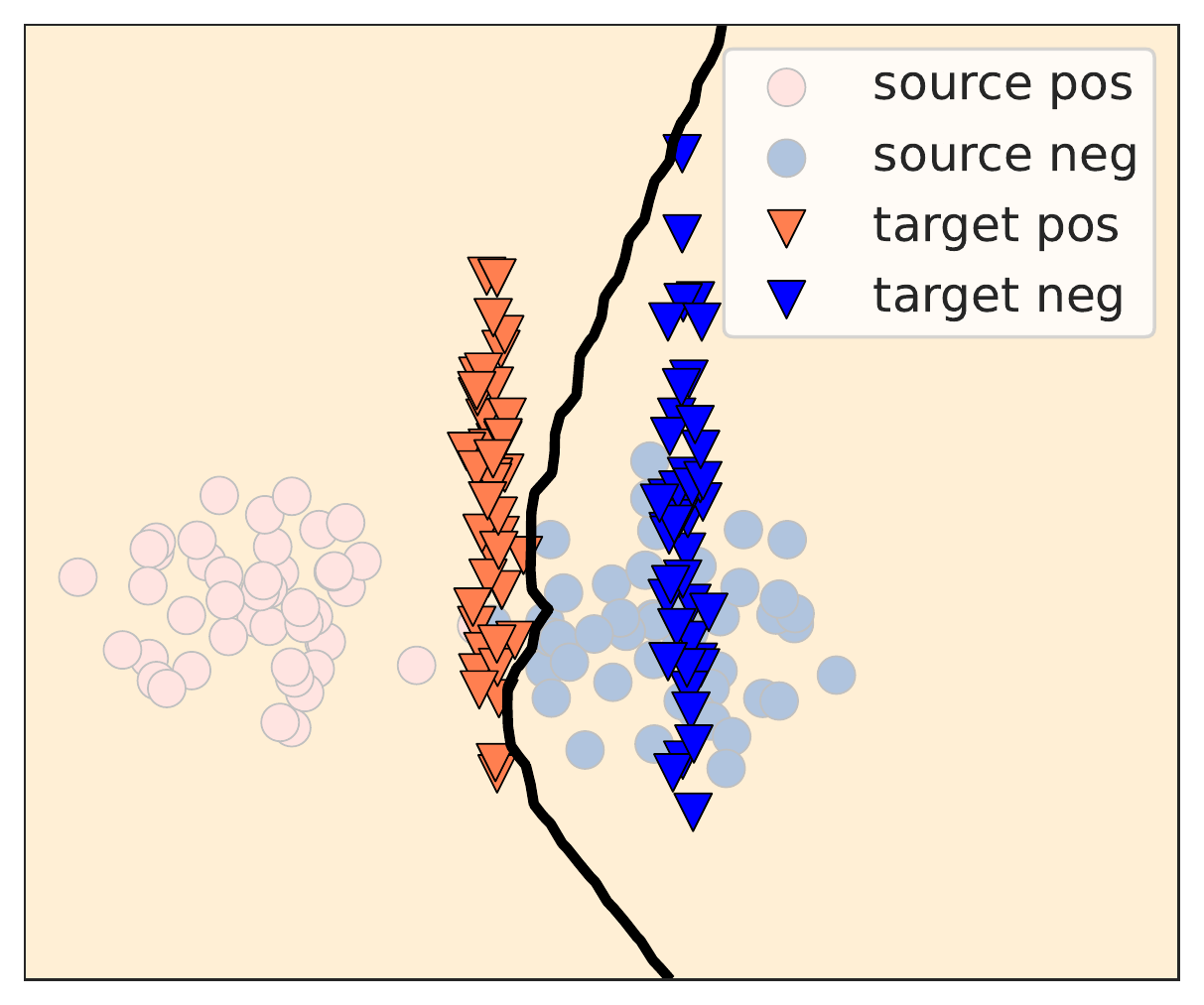}} \vskip-0.12in
  \caption{Example of EXPECTED optimized by PPS. (a) Pre-training on source data delivers the initially provided model. (b) The given model successfully adapts to target data through PPS within 80 queries.}
  \label{fig:toy}
\end{figure}

Fig.~\ref{fig:a} shows that the classifier is able to correctly classify two classes of source data after pre-training, but fails on target data. Fig.~\ref{fig:b} shows that the pre-trained classifier finally adapts to the target data successfully after conducting Alg.~\ref{alg1}. In particular, as most model parameters are frozen during tuning, we observe that the tuned model eventually maintains a good classification performance on source data as well.

%{\color{red}{split these into two paragraph, highlight the difference between Gaussian and reparamerterize}}

\subsection{Extension to Complex Models}
Typically, many tasks only tune a proportion of model parameters for better generalization (We actually have followed this idea in the example of Section~\ref{sec:toy_example}). In the rest of paper, we abuse $\theta$ as the tuned parameters and present how they are efficiently tuned when they are with a complex structure.

\subsubsection{Limitations of PPS and Our Strategy}
In many applications, $\theta$ consist of parameters distributed across different layers of DNNs. For example, researchers adjust all the batch normalization layers to narrow the source-target discrepancy~\cite{li2016revisiting,wang2020fully}. However, we point out that directly applying PPS in such a scenario suffers from two limitations: (1)~PPS is not sufficiently efficient if $|\theta|$ is large because the gradient estimation in a high-dimensional space is found less stable~\cite{li2019nattack}.
(2)~Treating $\theta$ as an entirety overlooks the fact that different layers should have different levels of impact~\cite{li2020few}, which means PPS may waste query chance on less contributive layers.

Our strategy for overcoming the above limitations consists of two techniques. \textbf{(1) Layerwise tuning.} \rt{To remedy the problem of inefficiency in updating all parameters~$\theta$ as a whole,} we partition the tuned parameters layerwise, i.e., $\theta = \{\ell_1,\ell_2,...,\ell_H\}$, where $\ell_h (1\le h\le H)$ represents the parameters of the $h$-th layer. We then propose to tune parameters $\theta$ more naturally; every time we only focus on updating a single layer's parameters $\ell_h$ while freezing the remaining layers, i.e., $\theta -\{\ell_h\}$. The basic idea is related to the sequential training on neural networks~\cite{belilovsky2019greedy}. However, they try to scale the end-to-end training to large size datasets while we focus on query-efficient tuning from feedbacks. \textbf{(2)~Query budget reassignment.} We model the importance of different layers by inspecting their performance improvements. Instead of leaning on the static weights from the prior knowledge~\cite{kirkpatrick2017overcoming}, we propose to dynamically assign more queries to the layers which receive bigger pay-offs.
%it is intuitively related to the study of layerwise attention~\cite{DBLP:conf/nips/HeTXHQ0L18}, but in tuning phase we do not alter the structure of base model.  we design an attention-like mechanism which assigns more query budget to more important layers. As a result, the 
\subsubsection{Query-Efficient Layerwise Tuning}
Let $\alpha \in \mathbb{R}^H$ be a layer importance vector. We intend to map it to a ($H-1$)-dimensional simplex, based on which a layer $l_h$ is sampled to be updated or not. As all the tuned parameters are deemed useful, a base probability $\frac{\gamma}{H}$ is maintained for each layer, and the sampling distribution $p$ is then written component-wise 
\begin{equation}\label{eq:pro}
    p_h = (1-\gamma)\frac{\text{exp}(\alpha_h)}{\sum_{i=1}^H \text{exp}{(\alpha_i})} + \frac{\gamma}{H}.
\end{equation}
In practice, we make the base probability deterministic to guarantee a least update for each layer. That is, we conduct a unit execution for every layer before samplings, which equals to give $u$ queries\footnote{As the number of parameters in different layers varies, $u$ is not identical for different layers in practice. We use the same $u$ for the convenient statement here.} to each layer beforehand. As the least update tells us which layers are more contributive, their respective performance improvements will be used to measure the layer importance. Specifically, for the $h$-th layer at the $t+1$-th iteration, the update rule for $\alpha$ is written as:
\begin{equation}\label{eq:alpha}
    \alpha_h^{t+1}=\alpha_h^{t}+\beta\underbrace{ \max \{0, \bar{E}_h^{t+1}(\theta)-\hat{E}_{h-1}^{t+1}(\theta)\} }_{\text{Average improvement}\; I_h^{t+1}},
\end{equation}
where $\beta$ represents how much we rely on the observed improvement from the least update. The involved average improvement $I_h^{t+1}$ is obtained by comparing with the last layer update. That is, $\bar{E}_h^{t+1}(\theta)$ denotes the average evaluation result of the candidates models that perturb $h$-th layer at the iteration of $t+1$, 
\begin{equation*}
    \bar{E}_h^{t+1}(\theta)=\mathbb{E}_{\ell_h \sim \mathcal{N}(\ell_h^t,\sigma^2)} E\left(\{\ell_1^{t+\frac{1}{2}},...,\ell_{h-1}^{t+\frac{1}{2}}, \ell_{h}^{t},..., \ell_{H}^{t}  \}\right),
\end{equation*}
and $\hat{E}_{h-1}^{t+1}(\theta)$ denotes the evaluation after $(h-1)$-th layer is updated by unit queries during the iteration of $t+1$,
\begin{equation*}
    \hat{E}_{h-1}^{t+1}(\theta)=E\left(\{\ell_1^{t+\frac{1}{2}},...,\ell_{h-1}^{t+\frac{1}{2}}, \ell_{h}^{t},..., \ell_{H}^{t}  \}\right).
\end{equation*}
Particularly, $\hat{E}_{h-1}^{t+1}(\theta)=\hat{E}_{H}^{t}(\theta)$ if $h=1$. 

Note that the above statement essentially suggests to split a single iteration into two stages. The first stage is in charge of the least update for every layer which yields the importance factors used for the queries reassignment during the second half. We emphasize that the first stage is indispensable because we want to inspect the response of every layer with the fact that only a few layers selected in the second stage. The complete algorithm, named Layerwise Coordinate Parameter Search (LCPS), is formally summarized into Alg.~\ref{alg2}. %In particular, we recycle the remaining queries if they are not run out during the reassignment (Steps~8-11). 
%We clarify three points about this algorithm. (1) For clearer statement, we introduce the auxiliary variable $I_k^{t+1}$ to denote the average improvement for $(t+1)$-th iteration (Step~4). (2) For each outer iteration, $\theta_k$ is at least updated once, denoted by $\theta_k^{t+\frac{1}{2}}$ (Step~3). (3) 

\begin{algorithm}[tb]
 	\small
 	\caption{Layerwise Coordinate Parameter Search (LCPS) }\label{alg2}
 	\begin{algorithmic}[1]
 	    \REQUIRE {Initially provided model $F_{\theta_0}$, query budget $Q$, learning rates $\eta$, $\beta$, batch size $b$, variance $\sigma^2$, unit size $u$}.
 		\FOR {$t=0,...,\lfloor Q/b \rfloor$}
 		     \FOR {$h=1,...,H$}
 		     \STATE  Update $\ell_h^{t+\frac{1}{2}}$ with $u$ queries following Alg.~\ref{alg1}.
 		     \STATE Compute average improvement $I_h^{t+1}$ through  Eq.~(\ref{eq:alpha}). 
 		     \STATE $\alpha_h^{t+1} \gets \alpha_h^{t} + \beta I_h^{t+1}$.
 		     \ENDFOR
 		     \STATE Compute $p^{t+1}$ by Eq.~(\ref{eq:pro}) with $\gamma=0$. 
 		     \FOR{$j=1,...,\lfloor(b-Hu)/u\rfloor$ } 
 	         \STATE	Sample a layer $h$ with $p_h^{t+1}$.
 	         \STATE Update $\ell_h^{t+1}$ with $u$ queries following Alg.~\ref{alg1}.
 		     \ENDFOR
        \ENDFOR
    \ENSURE  $\theta^{\lfloor Q/b\rfloor+1}$.
 	\end{algorithmic}
\end{algorithm}

\begin{figure*}[!t]
\centering
  \begin{minipage}[b]{0.32\textwidth}\label{fig:adult_performance}
    \centering
    \includegraphics[width=\textwidth]{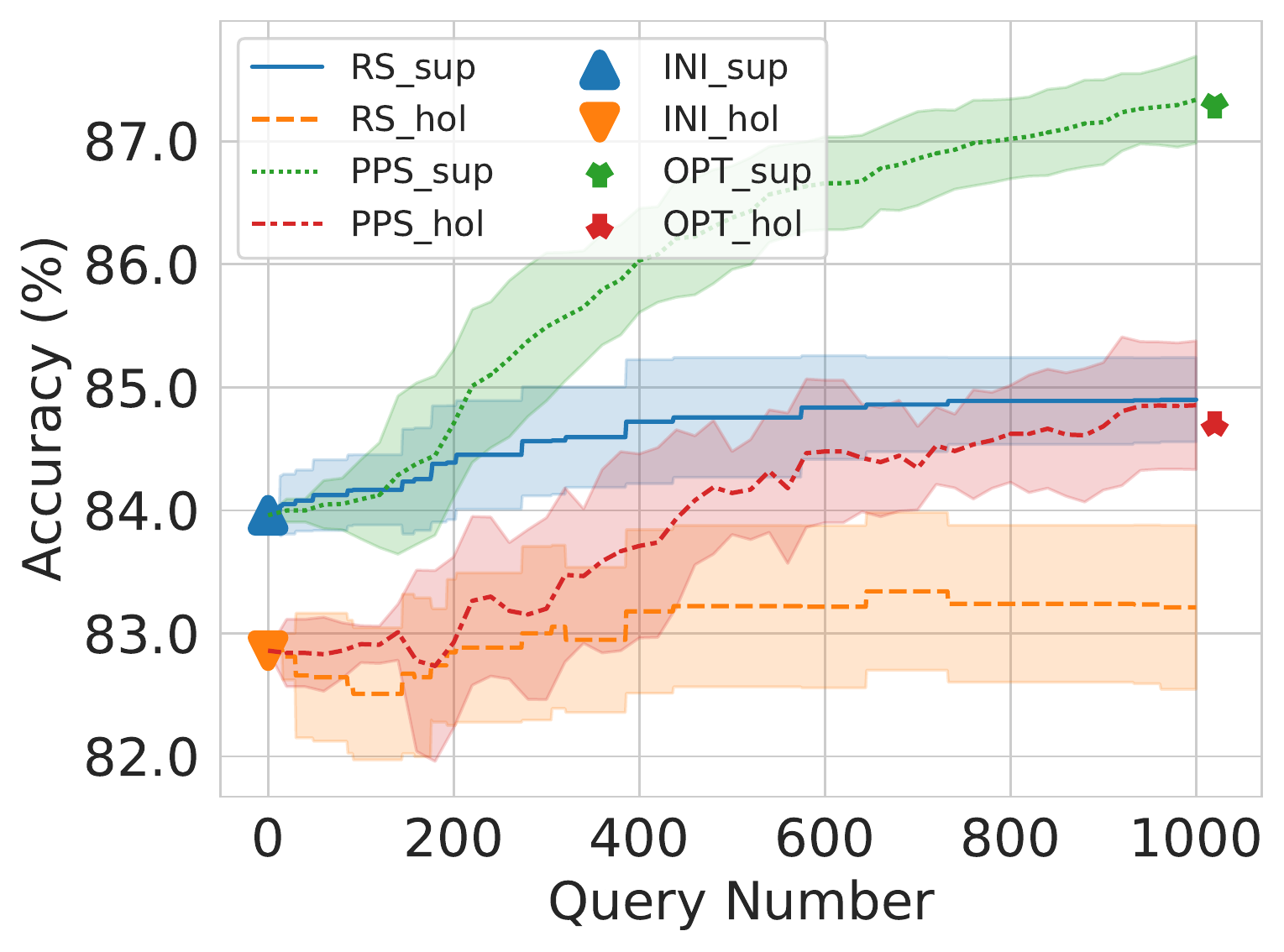}
    \centerline{(a) Adult}
  \end{minipage}
  \begin{minipage}[b]{0.32\textwidth}
    \centering
    \includegraphics[width=\textwidth]{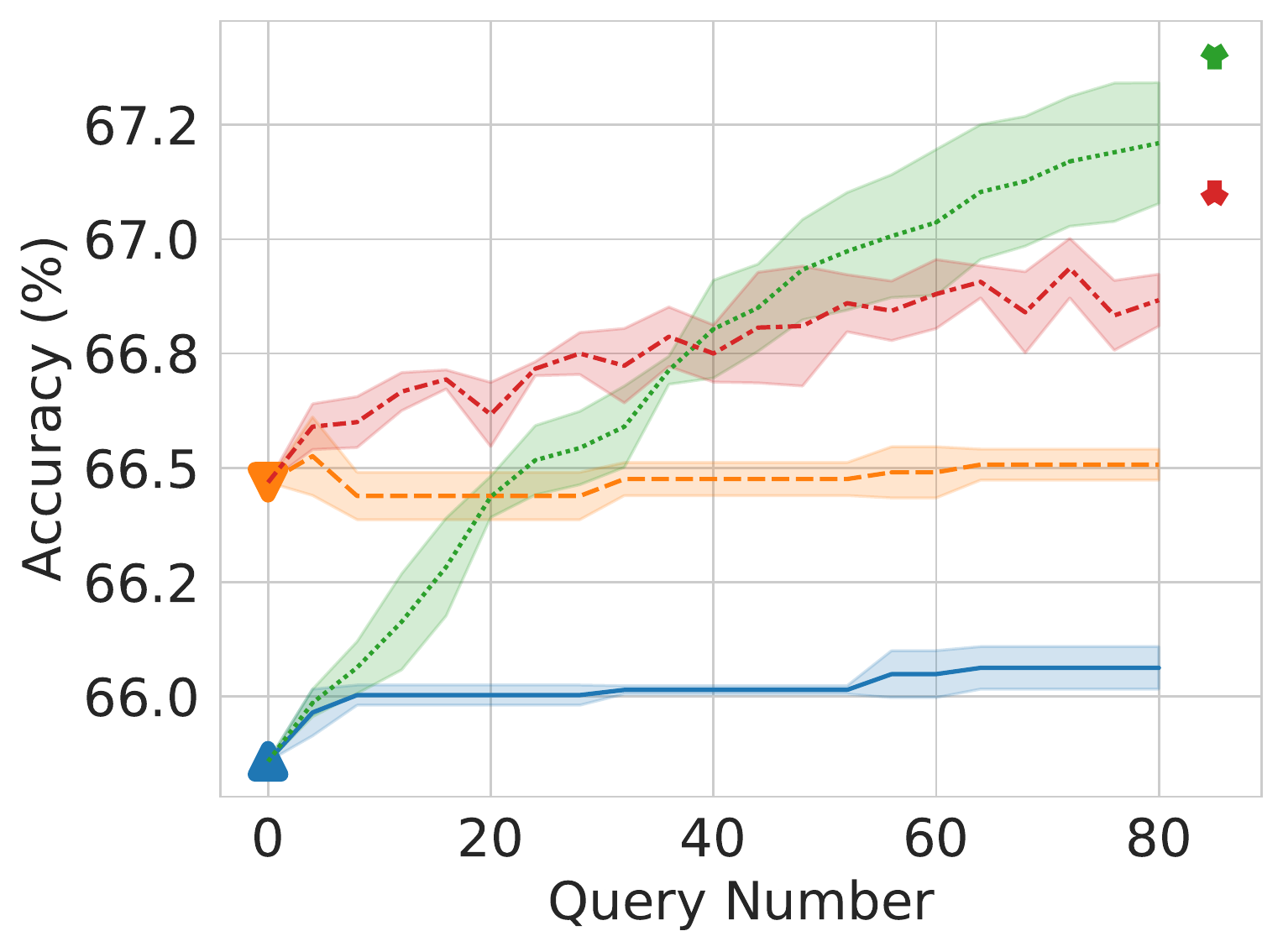}
    \centerline{(b) Amazon (good VOC)}
  \end{minipage}
  \begin{minipage}[b]{0.32\textwidth}
    \centering
    \includegraphics[width=\textwidth]{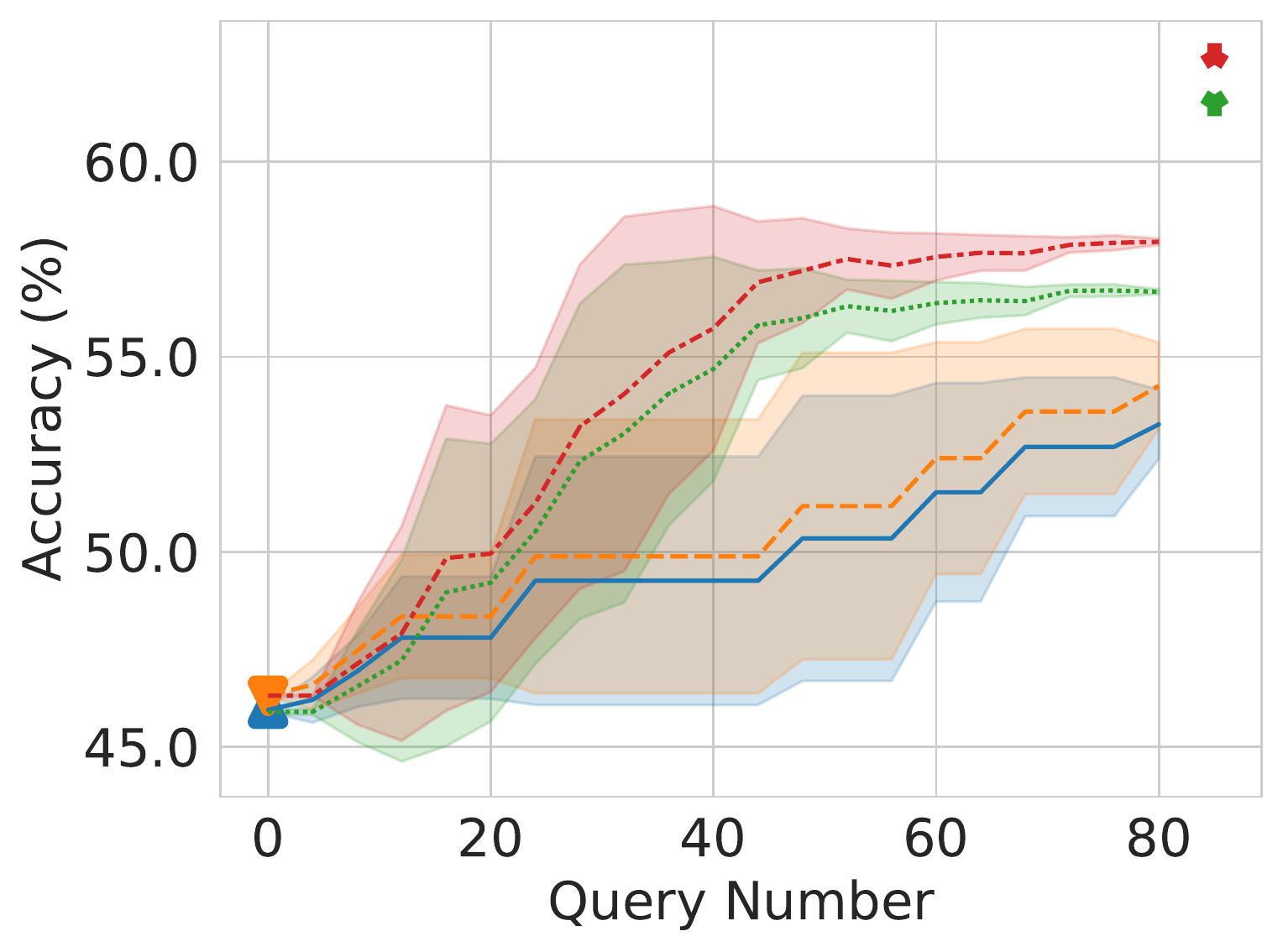}
    \centerline{(c) Amazon (bad VOC)}
  \end{minipage}
  \caption{Performance comparison on Adult and Amazon. Throughout all the experiments, the accuracy on the support set is monotonically non-decreasing, since we display the historically best at every iteration. Note that ``good VOC'' and ``bad VOC'' correspond to the different selections of vocabulary. The line shadow represents the standard deviation. \label{fig:exp_adult_amazon}}\vskip-0.2in
\end{figure*}

\subsubsection{Regret Analysis}
Reassigning different numbers of queries to different layers can be viewed as an exploration-exploiting game. Concretely, tuning without layer importance equals a pure exploration process, which is not a good option when the limited queries are given but layer discrimination does exist. By contrast, barely updating the most important layer corresponds to an exploiting strategy, which is not wise unless the query number is very small. If we regard tuning a specific layer as selecting a slot machine to play, our strategy can be understood to solve a multi-armed bandit problem~\cite{seldin2013evaluation}. In this sense, we aim to minimize the following expected regret,
\begin{equation}\label{eq:regret}
G_{\max}(T)-\mathbb{E}[G_{{\mathcal{A}}} (T)],
\end{equation}
where $T=1,2,...,\lfloor Q/b \rfloor$ is the horizon time, $G_{\max}(T)$ denotes the performance gain by picking the unknown optimal layer sequence, $G_{\mathcal{A}} (T)$ stands for the performance gain achieved by a designed algorithm, and the expectation is taken over the sampled layer sequences. 

Straightforward optimizing Eq.~\eqref{eq:regret} comes to an intractable problem, but this expected regret serves as a measurement to evaluate how the algorithm $\mathcal{A}$ approaches the oracle performance. Here we present the expected regret bound for the proposed LCPS.
\begin{theorem}\label{theorem2}
Given a deep model whose tuned parameters are $\theta = \{\ell_1,\ell_2,...,\ell_H\}$, for any $\beta >0$, we have that
\begin{equation*}
    G_{\max}-\mathbb{E}[G_{\text{LCPS}}] \le ({\beta}c(e-2)+1)G_{\max}+\frac{c}{\beta}\ln H
\end{equation*}
holds for any $T>0$, where $c = \frac{b-Hu}{u}$, ($b$ is the batch size, $u$ is the unit size), and $e$ is Euler's number. \hfill\rule{2mm}{2mm}
\end{theorem}
The proof of Theorem~\ref{theorem2} follows the sketch of Exp3 algorithm~\cite{seldin2013evaluation}. We leave it to Appendix for the readers who are interested in the differences. From Theorem~\ref{theorem2}, we can see that: (1) A smaller $c$ implies less expected regret, which suggests us to compute the query reassignment more frequently. However, we assume the performance gain hardly change in a few updates and thus attribute this problem to the selection of batch size $b$. (2) This weak regret bound is also a function of step-size $\beta$. From the Karush–Kuhn–Tucker (KKT) conditions, we obtain the regret reaches its minimum if we set $\beta = \sqrt{\frac{\ln H}{(e-2)G}}$ , where $G$ is the predicted maximum performance gain.
\begin{remark}
This regret bound is related to the scale of performance gain; the accumulative performance gain $G_{\max}$ could be very large if $T\rightarrow \infty$. Expect for some evaluation measurements like accuracy which naturally makes $G_{\max}$ upper bounded, one can do re-normalization to the immediate reward so that the bound of Theorem~\ref{theorem2} remains meaningful.
\end{remark}

\section{Experiment}
We remind readers two rules that all the experiments will obey. First, our experiments include how the initially provided model is produced, i.e., pre-training. But once the pre-training is completed, the source data is no longer used during the tuning, following the convention of standard model tuning. Second, we assume providers would only tune the parameters, which enables a lightweight modification on the users' side and also a small communication cost. 
% which corresponds to the experimental results on shifted data distribution (Section~\ref{sec:shifted_data_distribution}) and customized evaluation metric (Section~\ref{sec:beyond}). In particular, we explore the potential risk of EXPECTED when it is applied to defend target data from back-doors extraction~\cite{song2017machine} (Section~\ref{sec:private_tuning}), followed by the ablation study to the proposed two algorithms (Section~\ref{sec:ablation_study}).

%In this section, we empirically study the efficacy of our proposal. We start by verifying the generalization and query efficiency of our method. Then we explore some valuable properties of our method, such as flexibility and being extendable. %\footnote{Readers who are interested in model performance on standard datasets like ImageNet please refer to our supplementary for more details.}.

\subsection{Experimental Setup}
\textbf{Datasets.} Adult~\cite{kohavi1996scaling} is a tabular dataset for categorizing the annual income of different groups of citizens. As personal information is recorded, it is also a benchmark for fairness studies. Amazon review~\cite{mcauley2013hidden} is a text dataset that contains user comments about various products. The corrupted CIFAR-10/CIFAR-100~\cite{hendrycks2019benchmarking} are visual datasets where different type/level of corruptions simulate the real-world data noises. STS-B~\cite{cer2017semeval} predicts the semantic similarity between pairs of sentences which are extracted from different sources.%image captions, news headlines, and user forums. 

Except for the particular restatement, the main usages of the above datasets are described as follows. Adult has $14$ properties such as country, age, work class, education, etc, and we predict whether income exceeds $50K$ per year. We pre-train a binary classifier on the records with the country of ``U.S'' and take ``non-U.S'' records as unobserved target data. Amazon dataset is constructed from Amazon review data with two categories of products selected, i.e., ``Electronics'' and ``Watches''. In our experiments, the data-rich category ``Electronics" is used to pre-train a prediction model which maps user comments to the rating score ranging from one to five, and ``Watches" is treated as the target data. The settings of Adult and Amazon follows the work~\cite{chen2020comprehensive}. In terms of CIFAR-10-C/CIFAR-100-C, the initial provided model is built on clean images, and it is then tuned to fit the disjoint corrupted images following the unsupervised tuning research~\cite{wang2020fully}, which mimics the unexpected distribution shift in the real world. Last, we aim to tune BERT~\cite{devlin2018bert} and its variants on the STS-B task under EXPECTED. Following the research~\cite{hendrycks2020pretrained}, the models are firstly trained on the sentence pairs from the genre of MSRvid and then tuned to fit the unknown target data which are extracted from Images, where the evaluation metric is Pearson’s correlation coefficient. 

On the task of corrupted image classification, we treat all the corrupted images (target data) as the tuning data for a fair comparison with the unsupervised tuning methods. Throughout all the remaining datasets, the target data are split into two sets. We do randomly equal splitting for Adult and Amazon and use the default splitting for STS-B. \rt{One is the \emph{support set} that is used for evaluating the query efficiency of tuning algorithms, and the other is the \emph{holdout set} on which the model generalization is assessed.} The corresponding performances are denoted by ``sup'' and ``hol'', respectively.

\begin{figure*}[!t]
    \centering
    \begin{minipage}[b]{0.57\textwidth} %\vskip0.6in
     \begin{tabular}{lccc}
     \toprule
     Method & CIFAR-10-C & CIFAR-100-C & Note\\
     \midrule
     INI & $40.8$ & $67.2$ & - \\
     \midrule
     DAN~\cite{ganin2015unsupervised} & $18.3$ & $38.9$ & \multirow{4}{*}{Access $X$}\\
     TTT~\cite{sun2020test} & $17.5$ & $45.0$ \\
     BN~\cite{schneider2020improving}  & $17.3$ & $42.6$ \\
     Tent~\cite{wang2020fully}& $14.3$ & $37.3$ \\
     \midrule
     RS~\cite{bergstra2012random}  & $16.7$ & $40.5$ &$3K$ queries\\
     PPS & $15.1$ & $38.6$ & $3K$ queries\\
     LCPS& $13.8$ & $35.2$ & $1K$ queries\\
     \bottomrule 
     \end{tabular}\vskip0.15in
    \leftline{(a) Average error (\%) over 15 types of corruptions for the }
    \leftline{highest severity, where RS, PPS, LCPS, and Tent are based} \leftline{on test-time BN.}
  \end{minipage}
  \hspace{-0.5in}
  \begin{minipage}[b]{0.355\textwidth}
    \centering
    \includegraphics[width=\textwidth]{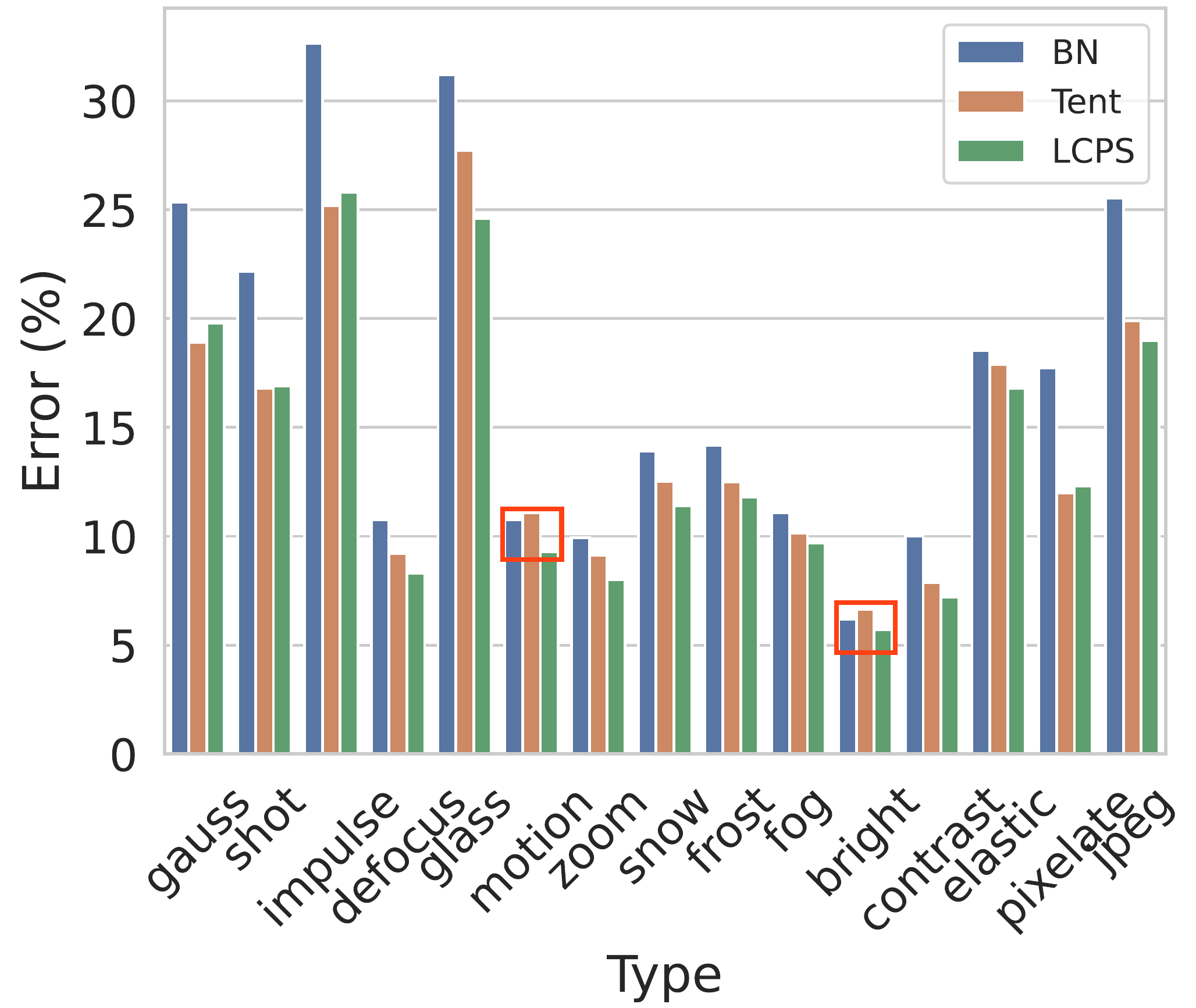}
    \centerline{(b) Error for each corruption on CIFAR-10-C. }
  \end{minipage}\vskip-0.1in
  \caption{\label{fig:cifar} (a) Comparison of different model tuning methods on CIFAR-10-C and CIFAR-100-C with the highest severity. (b) Performance of BN, Tent, and LCPS on CIFAR-10-C in terms of various corruptions.} \vskip-0.2in
\end{figure*}

\textbf{Models.} In respect of Adult, we use a 3-MLP with the penultimate layer of $80$ neurons. Following the fashion  of~\cite{kristiadi2020being}, we simply perturb the weights of the last layer which thus contains $80$ parameters for binary classification. For score prediction on Amazon, our implementation is based on the {\tenrm torchtext} library, in which the first layer is a mapping from the vocabulary to a latent representation, followed by three convolutional layers and a linear transformation to label space. The weights of the first layer are set as the tune parameters (with the size of $250K$) because the remaining layers are found less sensitive to the change of domains according to our experiments\footnote{Freezing partial parameters of the provided model is common but empirical in the model tuning community. Although our LCPS is able to tease out the useful parameters, this would consume a plenty of queries. In experiments, we would prefer employing LCPS only for ``certainly useful parameters".}. In terms of corrupted CIFAR-10/CIFAR-100, we use residual networks~\cite{he2016deep} with $26$ blocks which are implemented by {\tenrm pycls} library~\cite{radosavovic2019network}. We modulate features for target data by estimating normalization statistics and then update transformation parameters channel-wise. This setting is consistent with the recent research~\cite{sun2020test,liang2020we}, which turns out the tuned parameters make up a small proportion of the whole model. Similarly, we resort to tuning the layer normalization of BERT and its variants for predicting sentence similarity, managing to earn more performance improvement by tuning on target data.

\textbf{Baselines.} Naive baselines for EXPECTED include the initially provided model by pre-training and supervised tuning, which are dubbed as ``INI'' and ``OPT'', respectively. In addition, Random Search (RS)~\cite{bergstra2012random} is also considered here in a similar manner for hyperparameter tuning. The rest baselines are used for specific comparisons, whose results are retrieved from the literature or recomputed when it is possible. For the methods involving randomness, we repeat the experiments 10 times for a convincing comparison.

\subsection{EXPECTED on Shifted Data Distribution}\label{sec:shifted_data_distribution}
Distribution shift is one of the common roots for model tuning. This group of experiments justify the EXPECTED setting over different tasks under this configuration.

\textbf{Income classification.} We conduct Alg.~\ref{alg1} on Adult with the query budget $Q=1K$. Fig.~\ref{fig:exp_adult_amazon}(a) exhibits the classification performance on support set and holdout set, respectively. We observe that (1) PPS significantly improves the performance of INI and closely approaches OPT at $Q=1K$. (2) RS only achieves a subtle improvement to the initial model with the same number of queries and thus is less efficient for tuning model parameters.

\textbf{Rating prediction.} Alg.~\ref{alg1} is also run on Amazon, where a small query budget $Q=80$ is used\footnote{We empirically find the desired batch size on this task could be very small, and we use $b=4$ in our experiments. From Proposition~\ref{prop:1}, we suspect that the derived gradient for the tuning purpose lies in a very low dimensional space.}.  When a good vocabulary is carefully selected shown as Fig.~\ref{fig:exp_adult_amazon}(b), the improvement space w.r.t. the initially provided model is found quite limited, capped by the supervised tuning performance, i.e., OPT. In this case, the accuracy implemented by PPS only increases $1.3\%$ and $0.4\%$ on support set and holdout set respectively. By contrast, when a bad vocabulary is unintentionally selected shown as Fig.~\ref{fig:exp_adult_amazon}(c), we can see that PPS rapidly boosts the performance of INI, and it becomes stable after $50$ queries. In summary, the comparison between Fig.~\ref{fig:exp_adult_amazon}(b)\&(c) shows that (1)~PPS consistently earns more performance than RS regardless of different equipments of vocabulary. (2)~PPS is found sometimes trapped in a local optimum, which probably because the non-smooth evaluation function is insensitive to model perturbations. In other words, the multiple samplings with a fixed variance probably fail to help the model to escape from saddle points.

\textbf{Corrupted image classification.} We run both Algs.~\ref{alg1}\&\ref{alg2} on CIFAR-10-C/CIFAR-100-C with the query number of $3K/1K$ to tune batch normalization layers. In this experiment, more baselines are included, such as Domain Adversarial Netowrks (DAN)~\cite{ganin2015unsupervised}, Test-Time Training (TTT)~\cite{sun2020test}, test-time Batch Normalization (BN)~\cite{ioffe2015batch}, and test-time adaptation work Tent~\cite{wang2020fully}. Fig.~\ref{fig:cifar}(a) summarizes these methods and presents the average errors of tuning performance, where OPT with a supervised end-to-end tuning is omitted here because it can achieve a very low error. The results show that (1)~Tent is the most powerful among unsupervised tuning methods which access the entire features of target data, while the average error of LCPS is surprisingly better than Tent with $Q = 1K$. (2) Even offered more queries, RS and PPS cannot compete with LCPS, implying the advantage of Alg.~\ref{alg2} in tuning modern DNNs. Fig.~\ref{fig:cifar}(b) exhibits a close look to the specific results over each type of corruption. \ro{We observe that Tent updates towards the wrong direction on some particular corruptions, such as ``motion'' and ``bright'', causing even worse performance than BN. However, such performance drops do not happen to LCPS because feedbacks are simple but reliable (as label information is used during evaluations).}

\begin{figure*}[!bt]
\begin{minipage}[c]{0.3\textwidth}
        \vskip0.03in
        \includegraphics[width=\textwidth]{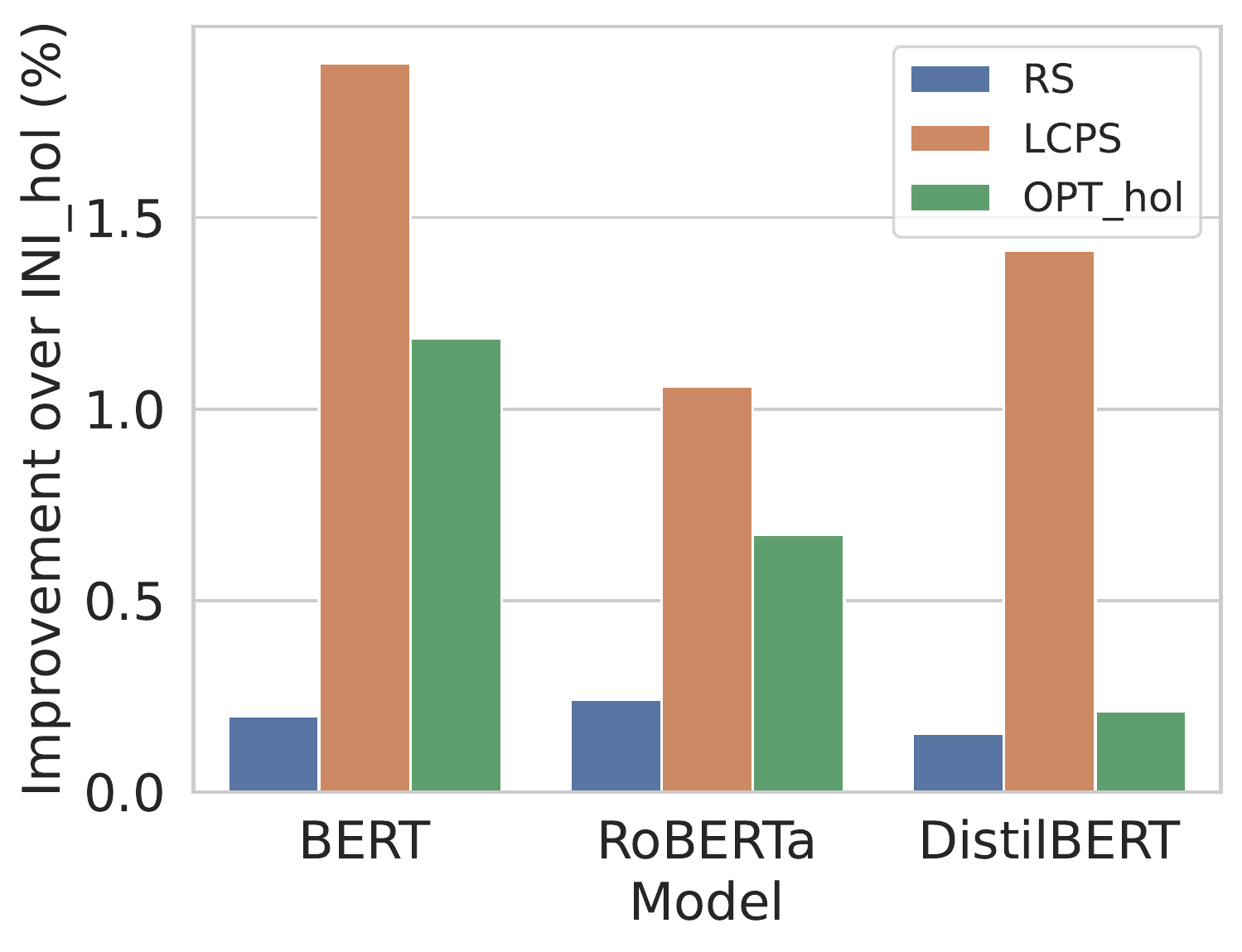}\vskip-0.05in
        \caption{\label{fig_sts} Generalization improvement of BERT and its variants after the model tuning on STS-B, which is computed by $\frac{s-s_0}{s_0}$, where $s_0$ and $s$ represent the model performance before and after tuning, respectively.}
\end{minipage}\hfill
\begin{minipage}[c]{0.33\textwidth}
        \includegraphics[width=\textwidth]{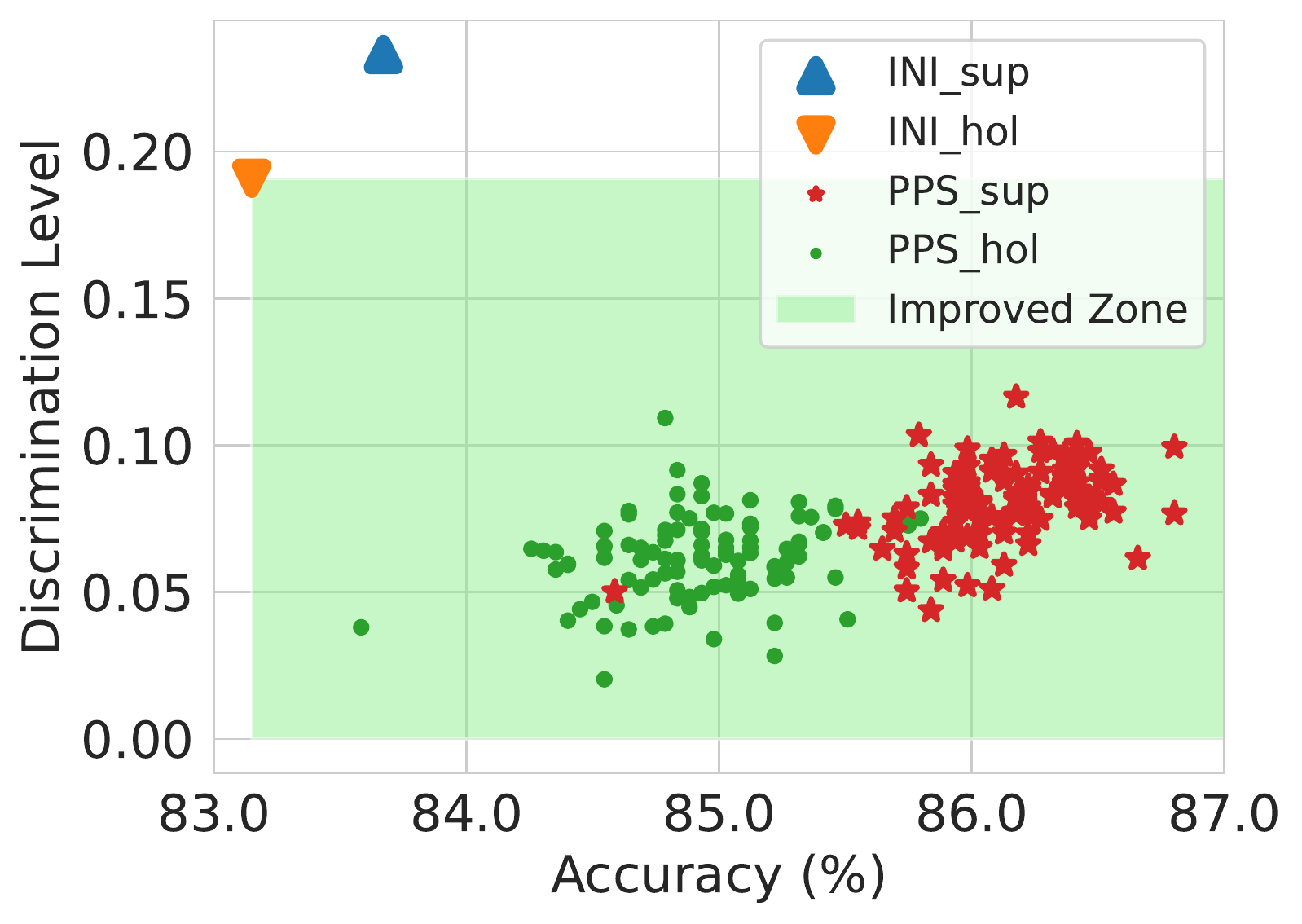}\vskip-0.1in
        \caption{\label{fig_acc_fair} Discrimination level reduction for model fairness tuning, where the particles falling in ``Improved Zone'' represent the models that have been improved in terms of both accuracy and fairness metrics on the holdout set.}
\end{minipage}\hfill
\begin{minipage}[c]{0.32\textwidth}
  \vskip-0.05in
  \caption{\label{table:topK}Evaluation performance (\%) of LCPS with top-$1$ or top-$5$ error as a tuning metric on two types of corruptions (Gaussian and Impulse noises) over CIFAR-10-C. ``Non'' represents an initially provided model with the test-time BN is directly evaluated without any tuning efforts. The lowest errors are marked as bold.}\vskip0.1in
  \centering
  \renewcommand{\arraystretch}{1.1}
  \setlength{\tabcolsep}{1.1mm}{	
  \scalebox{0.85}{
  \begin{tabular}{c|c|cc}
    \toprule
     \multirow{2}{*}{Type} & \multirow{2}{*}{Tuning} & \multicolumn{2}{c}{Error}\\
      &&Top-$1$ ($\downarrow$) & Top-$5$ ($\downarrow$) \\
    \midrule
    \multirow{3}{*}{Gaussian}& Non &$25.4$ & $3.0$ \\
    &w/ Top-$1$& {$\bm{17.7}^{\pm 0.13}$} & $1.5^{\pm 0.03}$ \\
    &w/ Top-$5$& $19.7^{\pm 0.18}$ & {$\bm{1.1}^{\pm 0.04}$}\\
    \midrule
    \multirow{3}{*}{Impulse} & Non &{$33.6$} & {$4.6$} \\ 
    &w/ Top-$1$ & {$\bm{21.9}^{\pm 0.09}$} & $2.1^{\pm 0.05}$  \\
    &w/ Top-$5$ &$24.6^{\pm 0.14}$ & {$\bm{1.4}^{\pm 0.03}$}  \\
    \bottomrule
  \end{tabular}}}%\vskip-0.07in
\end{minipage}
\end{figure*}

\textbf{Sentences similarity prediction.} Alg.~\ref{alg2} is also run on STS-B in which different models including BERT, RoBERTa, and DistilBERT~\cite{hendrycks2020pretrained} are examined. Pearson's correlation coefficient of each pre-trained model on the holdout set, i.e., the test set of Images, is $0.861$, $0.907$, and $0.849$, respectively. Again, we denote them by ``INI\_hol'' in terms of each backbone. Although these results are found comparable with what they behave on the source task~\cite{hendrycks2020pretrained}, we are interested to know whether they can be further tuned to achieve a better performance. By using $Q=5K$ queries, we apply RS and LCPS on these three models and the corresponding improvements are shown as Fig.~\ref{fig_sts}, where the standard tuning denoted by ``OPT\_hol'' is added in as a reference. The experimental results show that LCPS significantly improves the pre-trained model across different backbones, which certainly outperforms the RS strategy as well. Surprisingly, the standard tuning fails to upper bound LCPS as other experimental tasks. \ro{A reasonable explanation is that our LCPS utilizes \textbf{estimated full-batch gradients}, which is shown to better improve the generalization. Note standard tuning on BERT typically uses stochastic gradients which may be too noisy to provide effective information to update the model.}
%obtaining a full-batch gradient for standard tuning is a computationally intensive task that may not be feasible in practice.}

%the variance of stochastic gradients offsets the minor improvement while the whole-set performance feedbacks (summary statistics) are still useful in this case.

\subsection{EXPECTED for Customized Evaluation Metrics}\label{sec:beyond}
In some applications like the machine learning service provision, a customized evaluation metric might be needed for clients. Thus, the provided model which is never trained towards such an  objective usually cannot fulfill the downstream expectation. In this part, we study two interesting topics as the representatives of this situation. The first one is the fair classification where not only classification accuracy but also fairness critic is considered. The second one is fault-intolerant learning where the original evaluation metric is replaced by another metric in target tasks. For simplicity, we follow the basic configuration about datasets where the data distribution shift still exists, targeting a more challenging model tuning task.

\subsubsection{Fair Classification}\label{sec:fair_classification}
In this experiment, \emph{demographic parity}~\cite{hardt2016equality} is adopted as the fairness metric. Suppose a user requires a classifier which is unbiased on gender $z$ ($z=1$ denotes male and $z=0$ is for female) in terms of the high salary ($>\$50k$ per year). The corresponding discrimination level of demographic parity then can be defined by $\Gamma(\theta) = |\Pr(F_{\theta}=1|z=1)-\Pr(F_{\theta}=1|z=0)|$. That means every time after a local evaluation, the user will return a two-dimensional tuple with one element for the classification accuracy and the other for the discrimination level, i.e. $(E, \Gamma)$. Since two metrics commonly compete with each other~\cite{liu2022accuracy}, we propose to update their joint gradients as shown in Alg. 3 of Appendix.

Fig.~\ref{fig_acc_fair} shows the results of 100 independent executions of PPS under the above setting. Each green point denotes the model performance of a tuned model on the support set, and red stars are the corresponding performances on the holdout set. The most lower-right is the best. From this figure, we can see that (1) the particles falling in the green zone refer to the models which achieve improvements over the pre-trained model in terms of both classification accuracy and model fairness on the holdout set, making up $100\%$ of the whole trials. (2) The overall tuning accuracy is superior to testing while the discrimination level of testing is slightly better, implying an acceptable discrepancy between tuning and testing. (3) The discrimination level of INI has been dramatically decreased (by more than half) after tuning. Thus we can see our method under EXPECTED serves as an efficient fair-tuning approach for inaccessible data.

\subsubsection{Fault-intolerant Evaluation} 
One of the common fault-intolerance metrics is top-$K$ accuracy~\cite{chzhen2021set}. Unlike the single output prediction, top-$K$ classification produces lower errors. We take the multi-class classification task over CIFAR-10-C as an example. In our experiment, apart from tuning with top-$1$ error, top-$5$ error is used for tuning metric as a comparison. We achieve this by simply replacing the standard top-$1$ error with the top-$5$ error during the local evaluation. The experimental results on two corruption types with $2K$ queries, i.e., Gaussian and Impulse noise, are finally reported.

Fig.~\ref{table:topK} exhibits the results of tuning with top-$1$ and top-$5$ metric separately. When any of them is not used for tuning, its corresponding error is computed by evaluating the tuned model on the target task with this metric. For example, regarding images corrupted by Gaussian noises, the model tuned with top-$1$ metric under EXPECTED eventually achieves about $17.7\%$ error, and we also obtain its top-$5$ error by evaluating the tuned model with the top-$5$ metric whose performance turns out around $1.5\%$. From Fig.~\ref{table:topK}, we can see that (1)~our LCPS is efficient for the model tuning under EXPECTED because it has dramatically decreased the classification errors on both metrics. Notably, through $2K$ queries, the top-$1$ error has been decreased by around $7.7\%$ and $11.7\%$ on two types of corruptions, respectively. (2)~Although tuning with the top-$1$ metric decreases the top-$5$ error as well, the top-$5$ error could be reduced to a smaller value when it is directly used as the tuning metric. That means if the user demands a lower top-$5$ error, LCPS  naturally satisfies this requirement by straightforward replacing the top-$1$ error with top-$5$ error at the beginning of tuning. %(3)~It is worthwhile noticing that reaching a smaller top-$5$ error is with some sacrifice of top-$5$ accuracy. For example, for Gaussian noise corruption, the error decreases by $0.4\%$ on top-$5$ but its top-$1$ error increase by about $4.6\%$.

\textbf{Note.} The fairness metric is often hard to optimize since it is a group level measure defined on the entire tuning set. Top-$K$ error is non-differentiable which can be implemented by some extra operation like truncation. That means both of them need some elaborate design in a standard model tuning task. Interestingly, we emphasize that these metrics can be innocently used for our methods as we barely collect the emitted performance over them under EXPECTED.

\begin{figure*}[!t]
 \begin{minipage}[b]{0.32\textwidth}
    \centering
    \includegraphics[width=\textwidth]{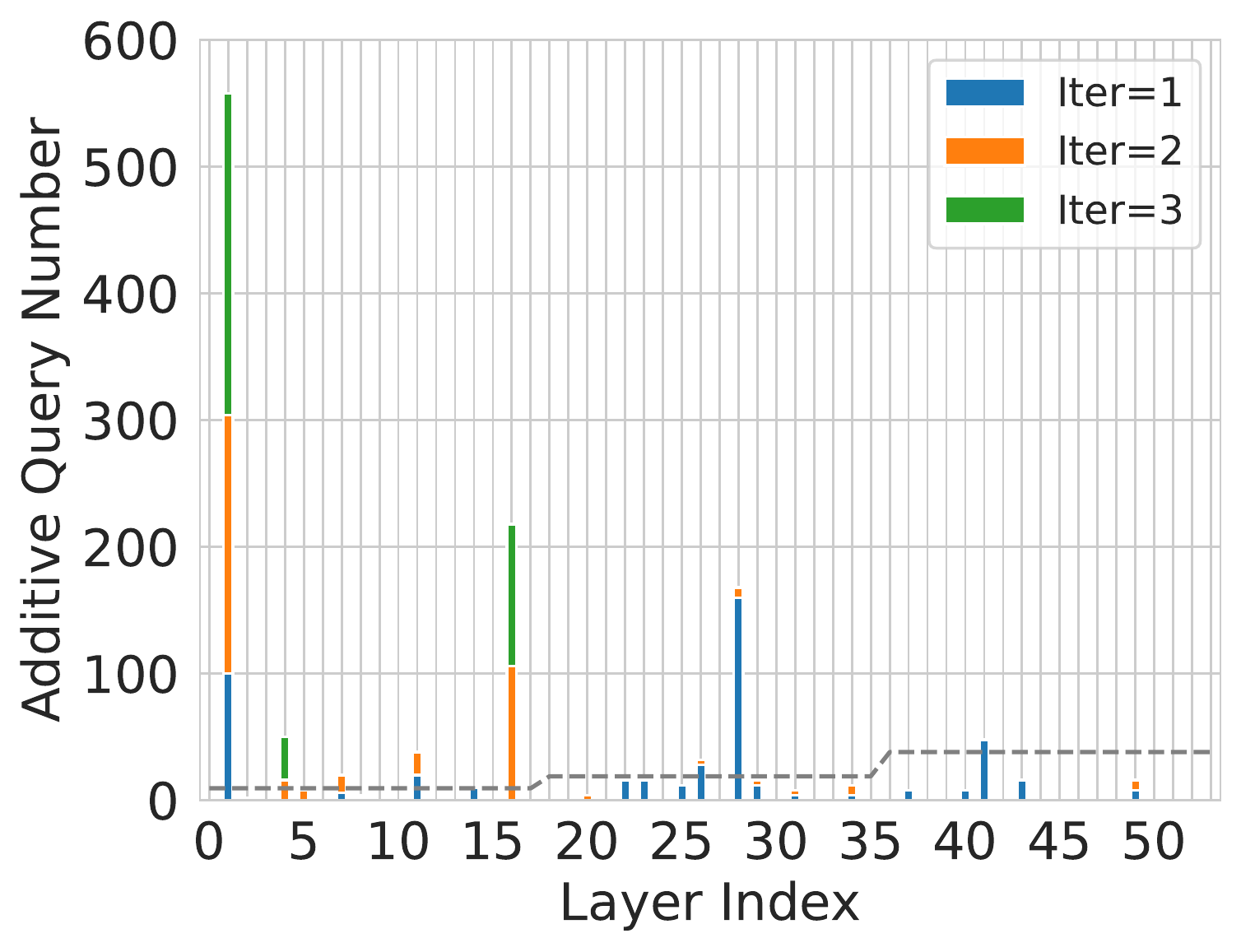}
    \centerline{(a) CIFAR-10-C}
  \end{minipage}
  \begin{minipage}[b]{0.32\textwidth}
    \centering
    \includegraphics[width=\textwidth]{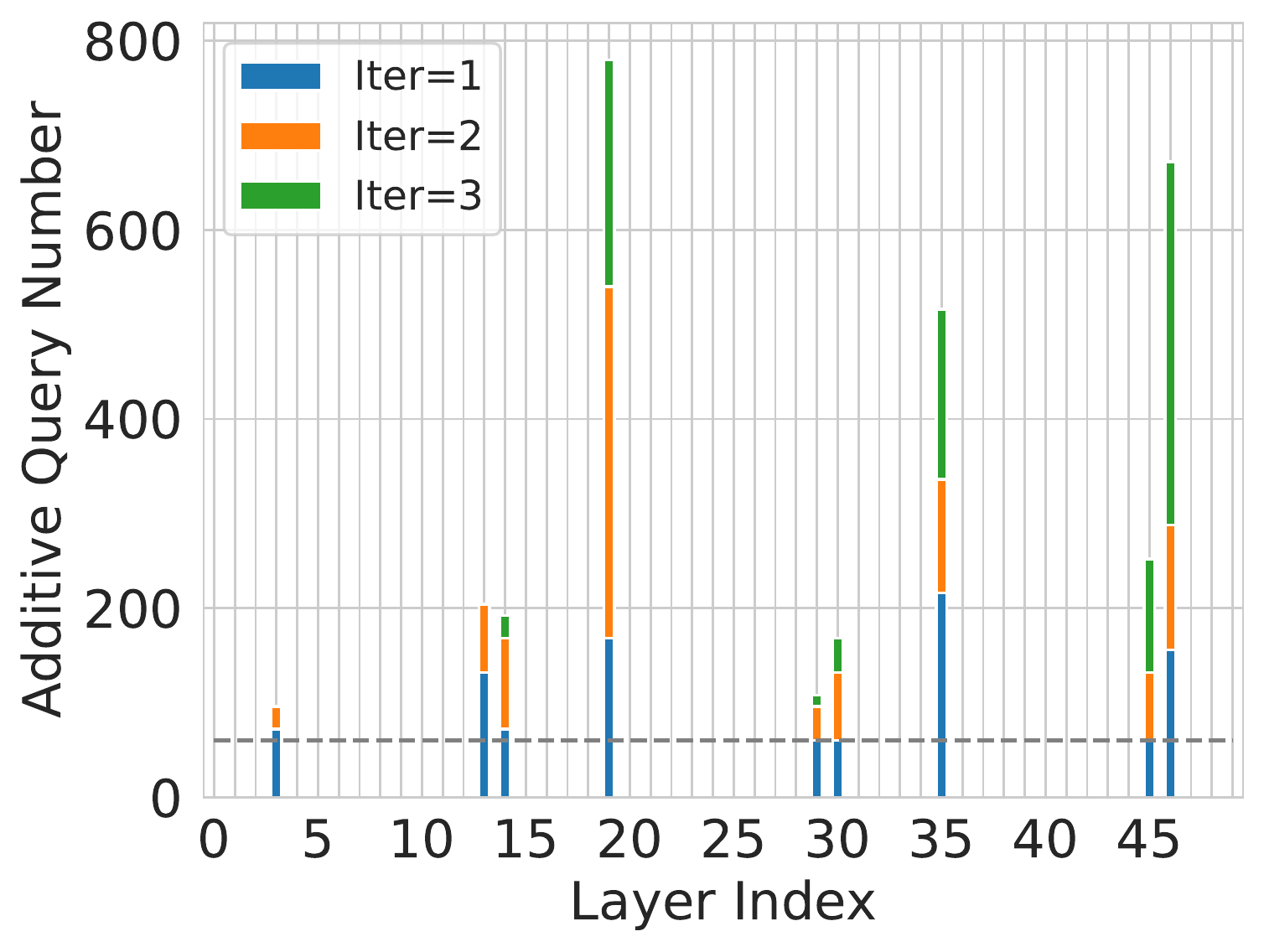}
   \centerline{(b) STS-B}
  \end{minipage}
  \begin{minipage}[b]{0.32\textwidth}
    \centering
    \includegraphics[width=\textwidth]{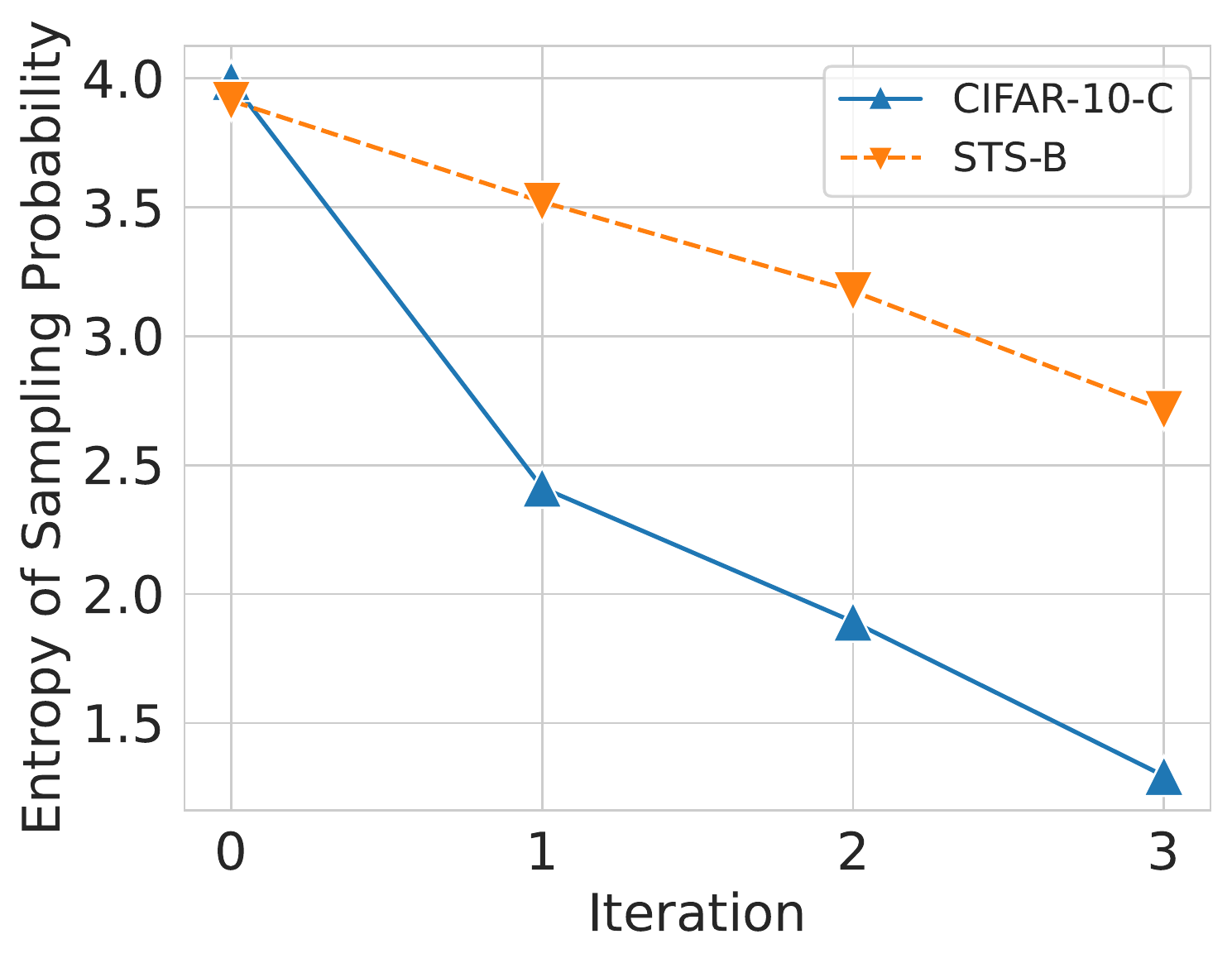}
    \centerline{(c) Entropy}
  \end{minipage}%\vskip-0.1in
  \caption{\label{fig:frequency} Query budget reassignment of LCPS on CIFAR-10-C and STS-B. (a) and (b) are corresponding the results of CIFAR-10-C with Gaussian corruptions and STS-B with BERT being backbones. The grey dashed line indicates the expected query assignment for each layer without the layer importance concern. (c) exhibits the entropy of sampling probability over each iteration for the two experiments.} %\vskip-0.15in
\end{figure*}

\begin{figure*}[!t]%\vskip-0.175in
 \begin{minipage}[b]{0.32\textwidth}
    \centering
    \includegraphics[width=\textwidth]{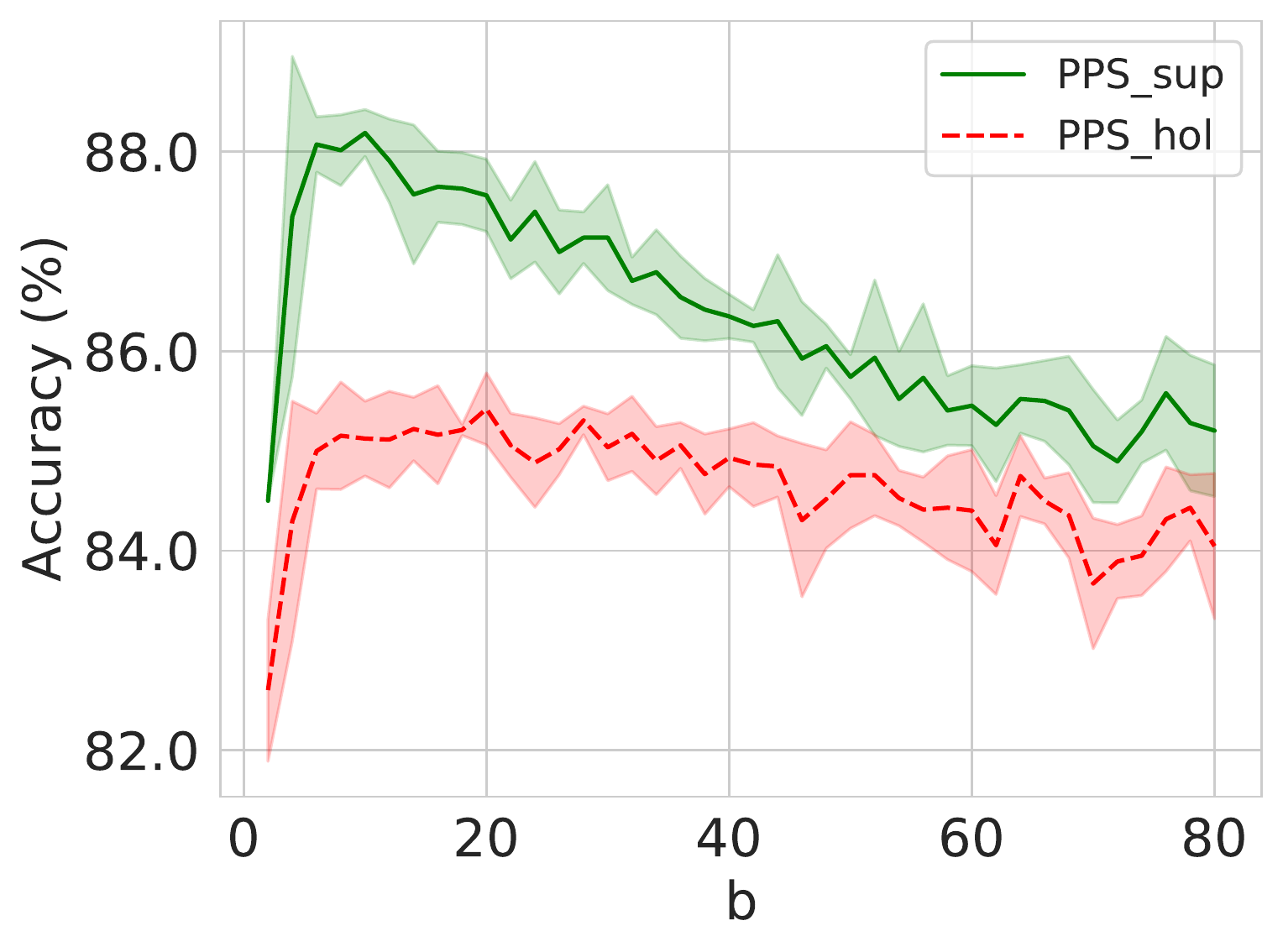}
    \centerline{(a) Sampling batch size}
  \end{minipage}
  \begin{minipage}[b]{0.32\textwidth}
    \centering
    \includegraphics[width=\textwidth]{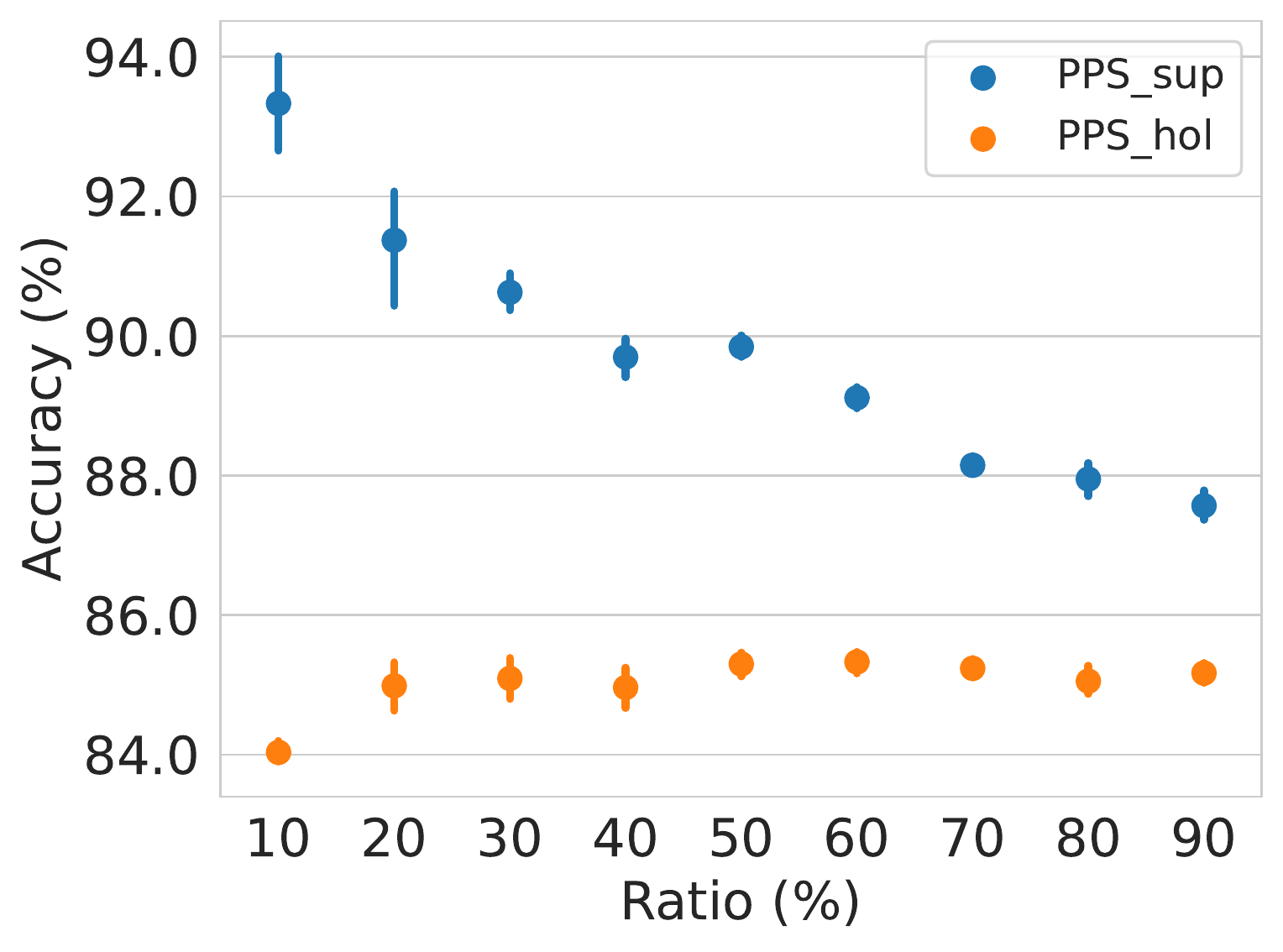}
   \centerline{(b) Support size}
  \end{minipage}
  \begin{minipage}[b]{0.325\textwidth}
    \centering
    \includegraphics[width=\textwidth]{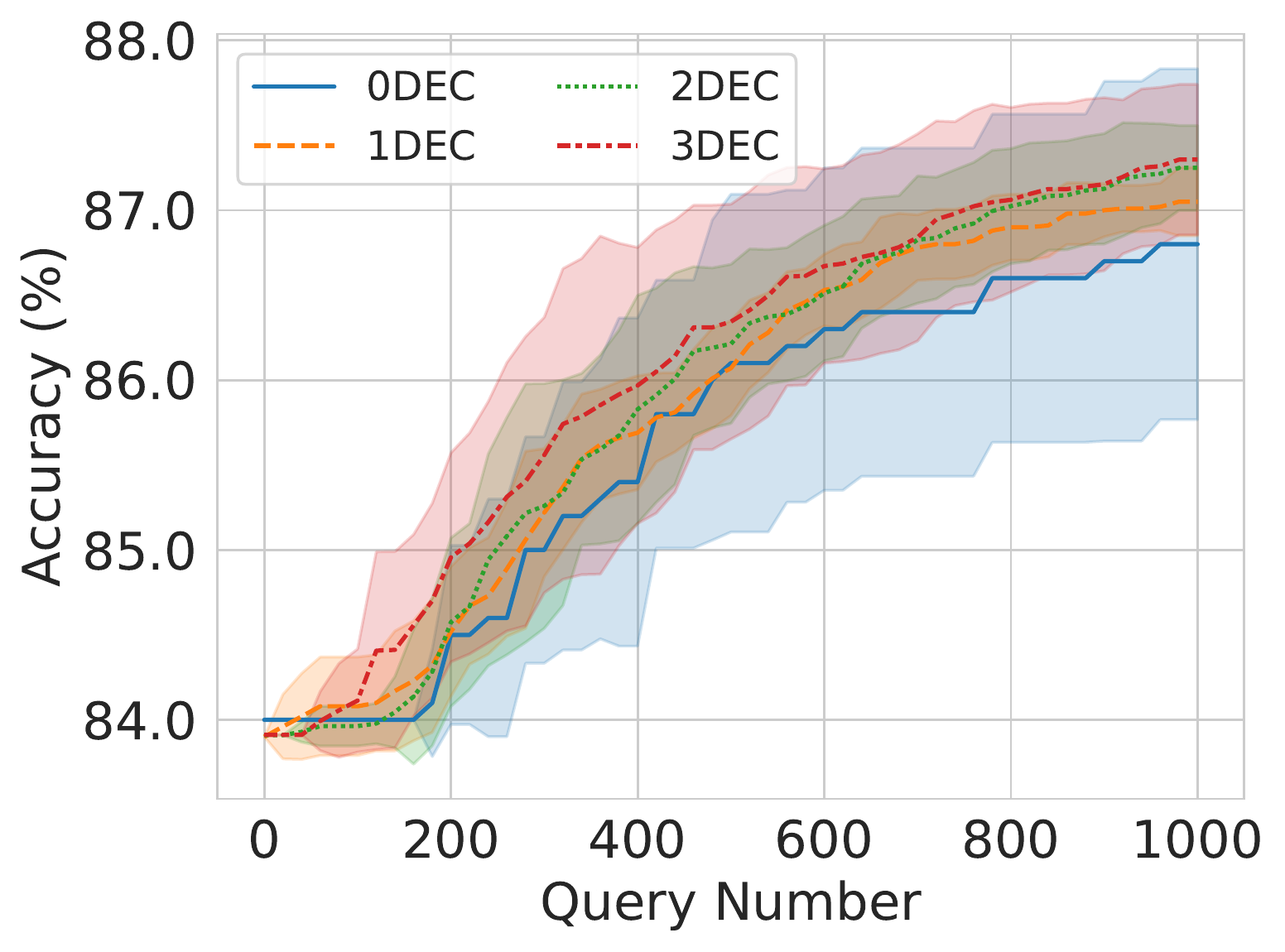}
    \centerline{(c) Precision of feedbacks}
  \end{minipage}%\vskip-0.1in
  \caption{\label{fig:ablation} Ablation study on three factors: sampling batch size, support size, and precision of feedbacks. ``XDEC'' in (c) means that the feedback value is rounded with X decimals.} %\vskip-0.15in
\end{figure*}

\subsection{A Close Investigation to LCPS}
We further investigate how LCPS works for complex models by visualizing the process of layerwise tuning on CIFAR-10-C (with Gaussian corruption) and STS-B (on BERT). The basic experimental settings follow Section~\ref{sec:shifted_data_distribution} but we let $Q=2K$ on CIFAR-10-C for a better comparison.

We present the experimental results in Fig.~\ref{fig:frequency}, which shows that (1)~in each iteration, only partial layers are selected in LCPS for updates, whose additive query numbers turn out much higher than the corresponding expected numbers (which are proportional to $|\ell_h| (h=1,2,...,H)$ and indicated by the grey dashed lines). (2)~A layer selected in the previous iterations would be prone to be selected again later. This is because the sampling probabilities of selected layers are much higher than the remaining ones due to the dominant average improvements (Refer to Eqs.~\eqref{eq:pro}\&\eqref{eq:alpha}). (3)~With the increment of iteration number, fewer layers are sampled, which means that the tuning process is towards exploitation given the limited query budget. This observation is also demonstrated by their steadily decreasing entropy of sampling probabilities (See
Fig.~\ref{fig:frequency}(c)). (4)~The additive queries which are reassigned during the second stage are dependent on the specific task. Roughly speaking, shallow features update is more crucial to CIFAR-10-C while STS-B prefers deep features. A possible explanation is that Gaussian corruption changes low features significantly while data genre in STS-B is encoded by some high-level information.

\subsection{Important Factors Study}\label{sec:ablation_study}
We empirically verify four factors that may have impact on the results.% and leave the discussion about other parameters to Appendix~\ref{appendix_params}.

\textbf{Sampling batch size.} \rt{We vary the sampling batch size from $2$ to $80$ with the step size of $2$ on Adult to investigate the trade-off between the precision of estimated gradients and the total number of update steps.
The results are shown as Fig.~\ref{fig:ablation}(a). In terms of the support set, the optimal performance is achieved when the batch size is about $10$. While it reaches the optima with the batch size being $20$ on the holdout set. Hence, we use $b=20$ throughout all other experiments on Adult. Related research~\cite{wierstra2014natural,hansen2016cma} suggests that $b$ is determined by the parameter dimensions, i.e., $b = 4+ \lfloor 3\log {|\theta|}\rfloor$. We find out this setup is useful for most cases except on Amazon. Recall that the sampling batch size for Amazon is quite smaller from Section~\ref{sec:shifted_data_distribution}. Therefore, we remind that this hyperparameter should be carefully selected, especially when the query efficiency is required. One possible workaround to this issue is resorting to an auxiliary validation set before executing tuning. }

%For other experiments, we set Therefore, tuning using our set-up batch size can generalize well on the holdout set. This verifies that our strategy to choose the batch size according to the performance on the support set is valid.

% Trader-off between batch size and update steps. Although PPS on support size prefers a smaller batch size we notice the generalization is steady over a ranger of. Without knowledge about batch size in practice, we use $b/4-b/2$.

\textbf{Support size.} We explore the effects of the size of support set by varying its ratio from $10\%$ to $90\%$ on Adult, and the results are shown as Fig.~\ref{fig:ablation}(b). With the increase of support size, it becomes harder to fit all the supported samples given the same query budget, but the model generalization, i.e., the accuracy on the holdset set, gradually improves. Additionally, we can see a smaller support set leads to a larger variance. By contrast, when more than $50\%$ of full support data ($>1000$ samples) is used, the model generalization becomes steady with a slight fluctuation only. This observation also suggests that EXPECTED does not demand a big support set, showing a desired trait for some data-scarcity scenarios. \rt{In practice, the support size should be increased to guarantee the generalization if the distribution shifts between the original pre-training data and target data are aware to be large, while it should not be decreased if collecting data is expensive.}

\textbf{Precision of feedbacks.} We study whether the precision of feedbacks has a direct impact in EXPECTED, which is also important when the back-doors attack~\cite{song2017machine} is concerned (Please refer to Appendix for more explanations). To this end, we run PPS on Adult by setting the number of decimals for the accuracy values from $0$ to $3$. The tuning performances on the support set are shown as Fig.~\ref{fig:ablation}(c), which demonstrates that (1)~zero decimal case fails to preserve the quality of feedbacks as the performance drops dramatically compared with the best configuration. (2)~The more precise feedbacks usually guarantee the better performance. However, as $\frac{1}{N_{\text{sup}}} > 0.01\%$ on Adult where $N_{\text{sup}}$ is the support size, $2$-decimal feedback is sufficient to use in this case. Hence, we can attribute the selection of the number of decimals to the side information about the support size.

\textbf{Layer importance.} To verify the necessity of developing LCPS for tuning complex models, we compare PPS and LCPS (with and without layer importance) through running them on CIFAR-10-C in terms of Gaussian and Impulse corruptions. Table~\ref{table:layer_importance} displays the corresponding results. We can see that LCPS only needs fewer queries to achieve the preset performances than both PPS and LCPS (w/o), showing a favourable property in tuning DNNs.

\begin{table}[t!]%\vskip-0.1in
  \caption{\label{table:layer_importance} The required query number ($K$) to achieve the preset tuning performance for two types of corruptions (Gaussian and Impulse) on CIFAR-10-C.}\vskip-0.07in
\centering
    \begin{tabular}{lccc}
        \toprule
         Type (\%)   & PPS & LCPS (w/o)  & LCPS (w/) \\
        \midrule 
         Gaussian ($22.0$) & $>10.0$ & $\sim 3.0$ & $\sim 0.2$\\
        Impulse ($20.0$) & - & $>8.4$ & $\sim3.5$\\
        \bottomrule
      \end{tabular}\vskip-0.2in
\end{table}

\section{Discussion}\label{sec:discuss}
We discuss the affinities of this work and existing research to clarify the scope of this study.

\textbf{Model tuning or model adaptation?} Our statement of not changing the semantic classes is consistent with the convention of domain adaptation~\cite{ganin2015unsupervised}. However, we use ``tuning'' instead of ``adaptation'' throughout this paper because of three reasons. (1)~Except some source-free studies~\cite{wang2020fully,liang2020we}, most domain adaptation works~\cite{quinonero2009dataset,ganin2015unsupervised,tzeng2015simultaneous,long2015learning} are doing the alignment between source and target data, while EXPECTED focuses on tuning a provided model to fit the target data only regardless of the performance on the source. (2)~The application of handling customized metrics on target data conceptually falls in the model tuning community, because the assumption of source-target distribution shift in domain adaptation is not a necessary requirement in the proposed setting. (3)~Standard tuning with the accessible target data usually upper bounds the proposed method. Technically, similar to the standard tuning, our methods can also apply to the case where semantic labels are changed and the classifier's head need renewing. Nevertheless, searching in such a huge space is more difficult to find the optimal solution, especially when a tight query budget is offered.

\textbf{Black-box optimization or reinforcement learning?} \ro{To the best of our knowledge, this is a first-of-a-kind work that conducts model tuning on inaccessible data through Black-box Optimization (BO). As a result, the PPS used for solving EXPECTED could be replaced by other alternative solutions in BO, such as CMA-ES~\cite{hansen2016cma} and Bayesian optimization\cite{snoek2012practical}. Note that we prefer PPS here because it achieves the close performance to CMA-ES and is scalable to higher-dimensional parameters as well (Refer to Fig.~\ref{fig:BO_comparison} and its experimental setup and result analyses in Appendix).} \rt{In addition, one may realize this challenge w.r.t. complex models is related to Reinforcement Learning (RL)~\cite{sutton1999policy} because we aim to find the optimal update strategy to maximize the accumulated reward (Eq.~\eqref{eq:regret}). Essentially, we cast layerwise tuning as a multi-armed bandit problem~\cite{seldin2013evaluation}, which is a said simple form of RL without \emph{state} modeling~\cite{sutton2018reinforcement}. We present more detailed analyses in Appendix for a clearer exhibition.}

%Trickily, we can understand the EXPECTED setting by ``putting data into a black box'' and requires model tuning design given only limited query chances (Recall the Unobserved Evaluation in Fig.~\ref{fig:Intro}(a))
\begin{figure}[!t]
    \begin{minipage}[a]{0.3\textwidth}
      \centering
      \includegraphics[width=\textwidth]{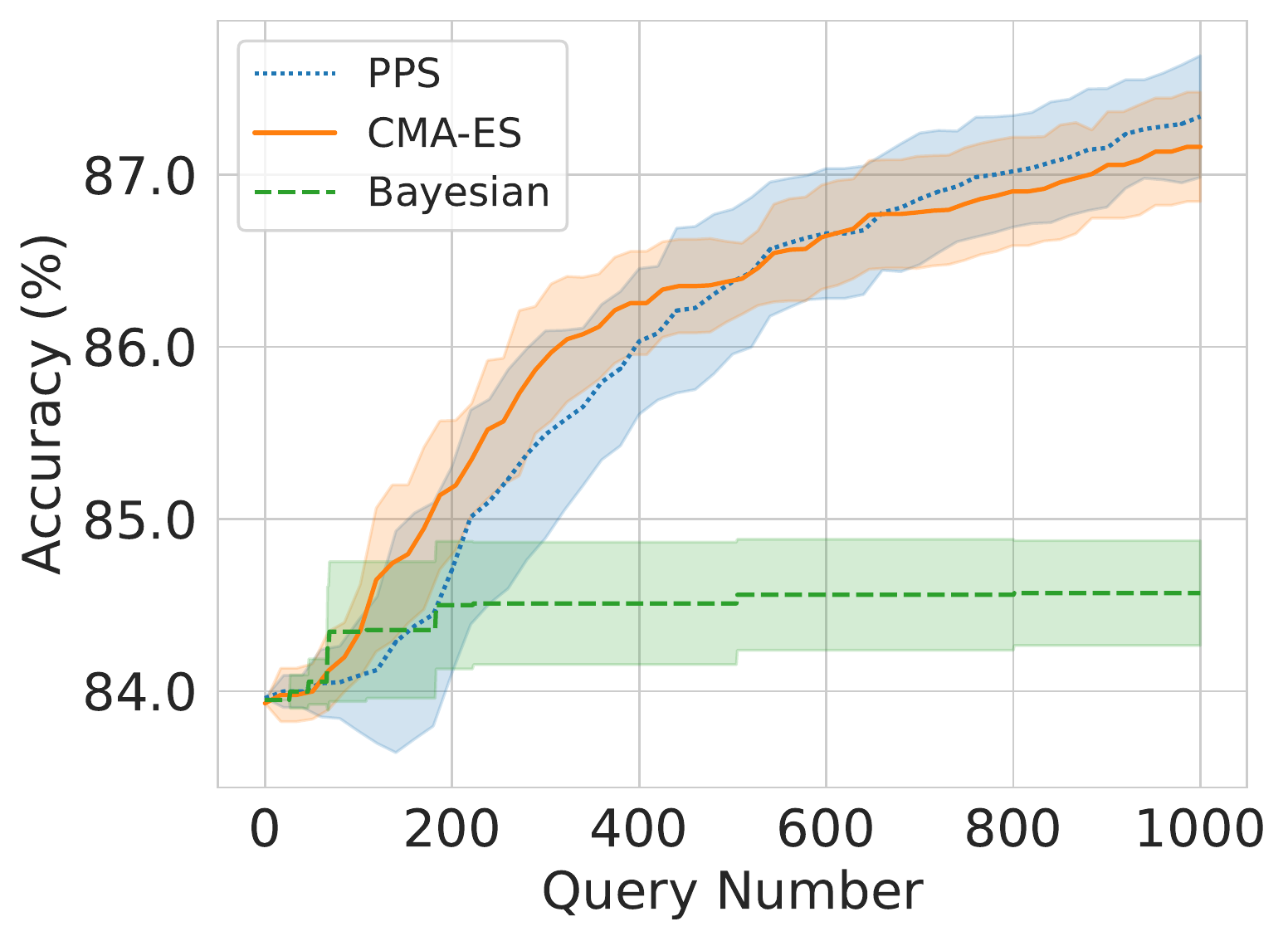}
    \end{minipage}
   %\hspace{0.1in}
   \begin{minipage}[b]{0.16\textwidth} 
   \renewcommand{\arraystretch}{1.0}
  \setlength{\tabcolsep}{1.0mm}{	
  \scalebox{0.8}{
     \begin{tabular}{lc}
     \toprule[1.3pt]
     Method & Time ($s$)\\
     \midrule
     \multicolumn{2}{c}{Adult ($|\theta| = 80$)}\\
     \midrule
     PPS & 2.7 \\
     CMA-ES& 2.9 \\
     Bayesian & 916.3 \\
     \midrule
    \multicolumn{2}{c}{Amazom ($|\theta| = 250K$)}\\
     \midrule
     PPS & $7\times10^3$ \\
     CMA-ES& N.A. \\
     Bayesian & N.A. \\
     \bottomrule[1.3pt]
     \end{tabular}}}
     %\vskip0.3in
  \end{minipage}%\vskip-0.1in
  \caption{\label{fig:BO_comparison} \ro{Comparison of three BO strategies. \textbf{Left}: tuning performance curves over query number on Adult. \textbf{Right}: running time (seconds) comparison on Adult and Amazon where the time cost of CMA-ES and Bayesian are not applicable due to OOM issues with high-dimensional parameters. }} 
\end{figure}

\section{Conclusion}
This paper presented a pioneer work of studying how to tune a provided model with only restrictive feedbacks on target task. To make this new setting clearer, we compared it with three summarized model tuning paradigms by carefully categorizing existing research. Our main technique borrowed the idea of NES~\cite{wierstra2014natural} to estimate the distribution of tuned parameters, in which we especially considered its practicability on tuning modern DNNs. The equipped theoretical analyses were to support the utility of the proposed methods.
Our experiments verified that the proposed methods can deal with the potential distribution shift (happening on target data) and customized metric problems. Besides, we revealed the properties of layerwise tuning strategy and explored some factors that may influence the experimental results.  

%results justify our setting and also show some potential applications, such as fair learning on private data, learning with indifferential metric, etc.  Model Generalization Improvement.

In the future, we will explore the following two directions. (1)~Query efficiency improvement. The specific form of Eq.~\eqref{eq:gradient} is actually not a unique choice for model parameters search. There are a number of surrogate objectives available from the literature~\cite{vu2017surrogate}. Recent study~\cite{turner2021bayesian} reveals that the ensemble of some surrogates achieves superior results for the hyperparameter tuning task. Inspired by this finding, we will explore if it is helpful to improve the query-efficiency in our methods. (2)~Extension to more general tasks. Some tuning applications may require modifying the model structures, e.g, in the multi-class classification, downstream data may have different semantic classes. Or model providers tries to change the model structure during tuning to expand more tuning space. Other extensions like tuning generative models are also interesting. For example, in molecular synthesis~\cite{guo2022data}, we cannot describe what the desired protein looks like, but there are multiple metrics for us to evaluate how good the protein-synthesis model is.

\bibliographystyle{IEEEtran}
\bibliography{reference}

\appendices
\section{Proof of Theorem 2}\label{appendix:proof_theorem}
We use Eq.~(x) to indicate the x-th indexed equation from the main paper, and use (x) for equation appearing in the Appendices.

\textbf{Theorem 2.} \textit{Given a deep model whose tuned parameters are $\theta = \{\ell_1,\ell_2,...,\ell_H\}$, for any $\beta >0$,
\begin{equation*}
    G_{\max}-\mathbb{E}[G_{\text{LCPS}}] \le \left({\beta}c(e-2)+1\right)G_{\max}+\frac{c}{\beta}\ln H,
\end{equation*}
holds for any $T>0$, where $c = \frac{b-Hu}{u}$, ($b$ is the batch size, $u$ is the unit size), and $e$ is Euler's number.}

\begin{proof}
Let $I^{t+1} \in \mathbb{R}^H$ denote a row vector whose $h$-th entry is $I_h^{t+1}$, and similar to $p^{t+1}$ and $p^{t}$. Then we have
\begin{align}
 &(I^{t+1})^T(p^{t+1}-p^t) \nonumber\\
 \overset{\circled{1}}{=}& (I^{t+1})^T\left(\text{softmax}(\alpha^t+\beta I^{t+1})-\text{softmax}(\alpha^t)\right) \nonumber\\
 \overset{\circled{2}}{\ge}& \beta (I^{t+1})^T \cdot \nabla_{\alpha^t}\text{softmax}(\alpha^t) \cdot I^{t+1} \nonumber\\
 \overset{\circled{3}}{\ge}&0, \label{eq:positive_dot_product}
\end{align}
where $\circled{1}$ follows Eq.~(11), $\circled{2}$ keeps only the first-order Taylor expansion, and $\circled{3}$ uses the fact that $\nabla_{\alpha^t}\text{softmax}(\alpha^t)$ is positive semi-definite~\cite{gao2017properties}. By rewriting ~\eqref{eq:positive_dot_product} into element-wise multiplication, we have the following inequality,
\begin{equation}\label{eq:first_moment}
    \sum_{h=1}^H p_h^t I_h^{t+1} \le \sum_{h=1}^ H p_{h}^{t+1} I_{h}^{t+1}.
\end{equation}
Suppose we are offered normalized average improvement $I_h^t \in [0,1]$, then we have
\begin{equation}\label{eq:second_moment}
    \sum_{h=1}^H p_h^t (I_h^{t+1})^2 \le 
    \sum_{h=1}^H p_{h}^{t} I_{h}^{t+1}.
\end{equation}

Let $W^t=\text{exp}(\alpha_1^t) + ...+\text{exp}(\alpha_H^t)$. By conducting LCPS, we have
\begin{align}
 \frac{W^{t+1}}{W^t} &= \sum_{h=1}^{H} \frac{\text{exp}(\alpha_h^{t+1})}{W^t} \nonumber\\
 &\overset{\circled{4}}{=}\sum_{h=1}^H \frac{\text{exp}(\alpha_h^t)\cdot \text{exp}{(\beta I_h^{t+1}}) }{ W^t }\nonumber \\
 &\overset{\circled{5}}{=} \sum_{h=1}^{H} p_h^t \cdot  \text{exp}(\beta I_h^{t+1})\nonumber\\
 &\overset{\circled{6}}{\le} \sum_{h=1}^{H} p_h^t [1+\beta I_h^{t+1} +(e-2)(\beta I_h^{t+1})^2]\nonumber\\
 &\le 1 + \beta \sum_{h=1}^H p_h^tI_h^{t+1} + (e-2)\beta^2 \sum_{h=1}^H p_h^t (I_h^{t+1})^2\nonumber\\
 &\overset{\circled{7}}{\le} 1 + \beta \sum_{h=1}^H p_h^{t+1}I_h^{t+1} + (e-2)\beta^2 \sum_{h=1}^H p_h^{t} I_h^{t+1}
\end{align}
where $\circled{4}$ follows Eq.~(12), $\circled{5}$ uses the definition of $p_h$ from Eq.~(11), $\circled{6}$ is derived from the inequality of $e^x \le 1+x+(e-2)x^2$, and $\circled{7}$ uses the \eqref{eq:first_moment} and~\eqref{eq:second_moment}.
Taking logarithms and using $1+x \le e^x$ comes
\begin{equation}\label{eq:wt_wt+1}
    \ln \frac{W^{t+1}}{W^t} \le \beta \sum_{h=1}^H p_h^{t+1}I_h^{t+1} + (e-2)\beta^2 \sum_{h=1}^H p_h^{t} I_h^{t+1}.
\end{equation}
Summing over $t$ of Eq.~\eqref{eq:wt_wt+1} we then have
\begin{equation}\label{eq:ww_le}
    \ln \frac{W^{T}}{W^0} \le \beta \sum_{t=1}^{T}\sum_{h=1}^H p_h^{t}I_h^{t} + (e-2)\beta^2 \sum_{t=1}^{T}\sum_{h=1}^H p_h^{t-1} I_h^{t}.
\end{equation}
Let $c=\frac{b-Hu}{u}$. By assuming that the average improvement $I_h^t$ stays constant in each batch optimization, we have
\begin{equation}\label{eq:exp_lcps}
    \mathbb{E}[G_{\text{LCPS}}] = \sum_{t=1}^T \left( \sum_{h=1}^H I_h^t + c \sum_{h=1}^H p_h^t I_h^t \right).
\end{equation}
Taking Eq.~\eqref{eq:exp_lcps} into Eq.~\eqref{eq:ww_le}, we get
\begin{equation}\label{eq:ww_le_exp_lcps}
\begin{aligned}
    \ln \frac{W^{T}}{W^0} \le
    \frac{\beta }{c}
    &\left( \mathbb{E}[G_{\text{LCPS}}] -
    \sum_{t=1}^{T}\sum_{h=1}^H I_h^{t} \right) \\
    &+ (e-2)\beta^2 \sum_{t=1}^{T}\sum_{h=1}^H p_h^{t-1} I_h^{t}.
    \end{aligned}
\end{equation}
For any layer $j$ is selected,
\begin{equation}\label{eq:ww_ge}
   \ln \frac{W^{T}}{W^0} \ge \ln \frac{\text{exp}{(\alpha_j^{T})}}{W^0} = \beta \sum_{t=1}^{T} I_j^{t}-\ln H.
\end{equation}
Combining Eqs.~\eqref{eq:ww_le_exp_lcps} and~\eqref{eq:ww_ge}, we arrive at
\begin{equation}\label{eq:exp_lcps_ge}
\begin{aligned}
 \mathbb{E}[G_{\text{LCPS}}] \ge
 \sum_{t=1}^{T}\sum_{h=1}^H I_h^{t}
 &+ c\sum_{t=1}^T I_j^t
 - \frac{c}{\beta } \ln H\\
 &-\beta c(e-2) \sum_{t=1}^{T}\sum_{h=1}^H p_h^{t-1} I_h^{t}.
\end{aligned}
\end{equation}
In addition, we can verify that
\begin{equation}\label{eq:G_max}
    \sum_{t=1}^{T}\sum_{h=1}^H p_h^{t-1} I_h^{t}
    \le \sum_{t=1}^T \max_h I_h^t \le G_{\max}.
\end{equation}
Combining Eqs.~\eqref{eq:G_max} and~\eqref{eq:exp_lcps_ge} lets us obtain the inequality of the Theorem~2.
\end{proof}

\section{Algorithm 3 for Fairness Learning}
\label{appendix_alg_fair}
In the fair classification task, we refine an initially provided model relying on both the model accuracy and fairness. Demographic parity is used as the fairness metric in this experiment. By employing an extra weight factor $\rho$ to balance two measurements, we update the model by Alg.~\ref{alg_fair}. $\rho=0.4$ is used for the reported experimental results in Fig.~6.
     \begin{algorithm}[H]
        \caption{Performance-guided Parameter Search (PPS) for Fair Classification}\label{alg_fair}
        \begin{algorithmic}
          \REQUIRE {Initially provided model $F_{\theta_0}$, query budget~$Q$, learning rate $\eta$, batch size~$b$, variance~$\sigma^2$, weight factor $\rho$}
          \FOR {$t=0,..., \lfloor Q/b\rfloor$}
          \STATE Sample $\{\epsilon_j\}_{j=1}^{b/2} \sim \mathcal{N}(0,I)$, and for each $j$ get $\epsilon_{b-j+1} = -\epsilon_j$.
          \STATE Generate candidate models $\{\theta_i\}_i^b$ as queries where $\theta_i = \theta^t+\sigma\epsilon_i$.
          \STATE Collect and normalize $\{E(\mathcal{D};\theta_i), \Gamma(\mathcal{D};\theta_i)\}_{i=1}^b$.
          \STATE $\theta^{t+1} \gets \theta^{t}-\frac{\eta}{\sigma b}\sum_{i=1}^b \epsilon_i[\rho E (\mathcal{D};\theta^t+\sigma\delta_i)-\Gamma(\mathcal{D};\theta^t+\sigma\delta_i)]$
          \ENDFOR
          \ENSURE $\theta^{\lfloor Q/b\rfloor+1}$
        \end{algorithmic}
      \end{algorithm}
\section{Discussion of Private Model Tuning }\label{appendix:private_tuning}
\begin{figure}[!ht]
 \centering
  \includegraphics[width=7cm]{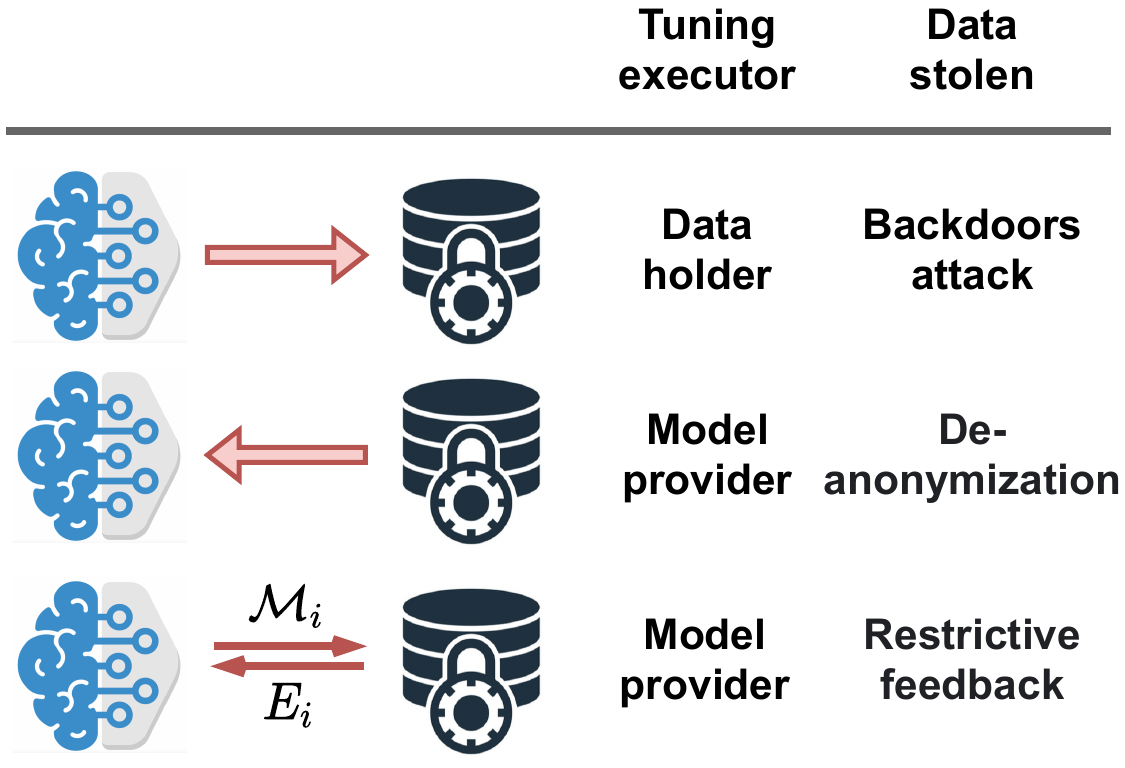}
  \caption{\label{two-way_private_fine_tuning} Comparison among three forms of model tuning. }
  %Top row: data holder receives a pre-trained model and executes model tuning in an unobserved manner. Middle row: data holder pre-processes local data (e.g., anonymization) and send them to model provider for tuning. Bottom row: every time model provider slightly modifies the pre-trained model and send it to data holder as a query, and data holder returns its performance on private data as a feedback. This interaction does not terminate until the model performance is satisfactory or queries run out.
\end{figure}

Preventing the adversarial inference about tuning data during the model tuning is of significance when target data involves sensitive information. We present two instances shown as the first two rows in Fig.~\ref{two-way_private_fine_tuning} which illustrate how existing works conduct model tuning with the data privacy concern. 

\emph{One-way data holder tuning.} Given a pre-trained model (e.g. downloaded from the internet under some agreed licenses), a data holder can execute model tuning on the local private data in an unobserved manner. Once the model is accessible in a white-box or black-box way after tuning, this one-way data holder tuning is at the risk of data leakage if the model provider provide a back-door model to data holder~\cite{song2017machine}. For example, private data could be encoded in the least significant (lower) bits of the deep model's parameters (white-box) or the label vector of augmented data (black-box) to intentionally extract private data. Please also note in this case, the original source model should be stored on local device and the data holder is assumed to be capable to model tuning.

\emph{One-way model provider tuning.} Alternatively, a data holder can pay experienced model providers to do tuning on the model providers' side. In this case, to preserve data privacy, the main attention of the data holder is on how to ``reedit'' private information before sending them to a model provider. Anonymization~\cite{zhou2008brief} seems a workaround to this problem, but it is quite limited to tabular data, and it has been demonstrated weak by de-anonimization~\cite{porter2008identified}. Other techniques like local differential privacy~\cite{cormode2018privacy} by randomizing raw feature is of low utility for real world applications.

Essentially, the above two instances do not change the tuning process itself; just like training, both of them straightforwardly feed the (original/edited) target data to the model for update. We realize the common root of data leakage for two instances is that they allow the model and data to stay on the same side, which serves as the base of the standard model tuning. As the goal of model tuning is to deliver a satisfactory model for data holders, we find that the introduced EXPECTED is a solution for this challenge. %a simple and rational option is to return the model performance as feedback to model provide. 

%This inspires us to think about a question -- can model provider optimize the pre-trained model without peeking on private data? The answer is YES, because from the perspective of machine learning, we are able to update model by data statistics instead of raw data.  

\emph{Two-way EXPECTED.} EXPECTED keeps data and model staying on their sides. Without any demands to data holders' ability on model tuning or data edition like previous two instances, EXPECTED only requires data holders to do the model evaluation on private data and return the performance to the model provider. Within limited queries, the model provider is expected to craft a satisfactory model for data holder. The key here is that model providers only receive restrictive feedbacks which will not expose much information about the private data. For example, in case data information is encoded via feedback scores, we have shown in Section~5.5 that it will not be very risky as $2$ decimals precision might be sufficient to use in EXPECTED. 

\rt{\emph{Computation cost and efficiency.} Given a query budget $Q$, query model $F_{\theta}$, the inference cost on support set $m$, and the feedback vector $s$, the \underline{communication cost} is $Q(|F_{\theta}|+|s|)$ and the \underline{total inference cost} amounts to $Qm$. As every $b$ queries contribute one gradient, according to Eq.~(9) in the main manuscript, the \underline{model tuning cost} can be written as $\frac{Q|\theta|}{b}$. All three costs are controllable via the query budget $Q$. The efficiency in this work refers to the query efficiency, which is mainly verify that our methods can achieve the same performance with baselines but using fewer queries.}

\section{\rt{Two Real Scenarios}}
\rt{We provide two real scenarios to highgith the significance of EXPECTED.}

\rt{\emph{Real case I.} Tuning foundation models for downstream tasks. Training large foundation models (e.g., GPT-3~\cite{brown2020language}) is expensive, and thus trained models are usually proprietary and not made public. Meanwhile,  downstream users may not be inclined to share their private data with model owners because of privacy concerns. Given two-way communications, our EXPECTED setting enables tuning a personalized/customized model for downstream users while not peeking at their data.}

\begin{figure}[h]
    \centering
    \includegraphics[width=0.47\textwidth]{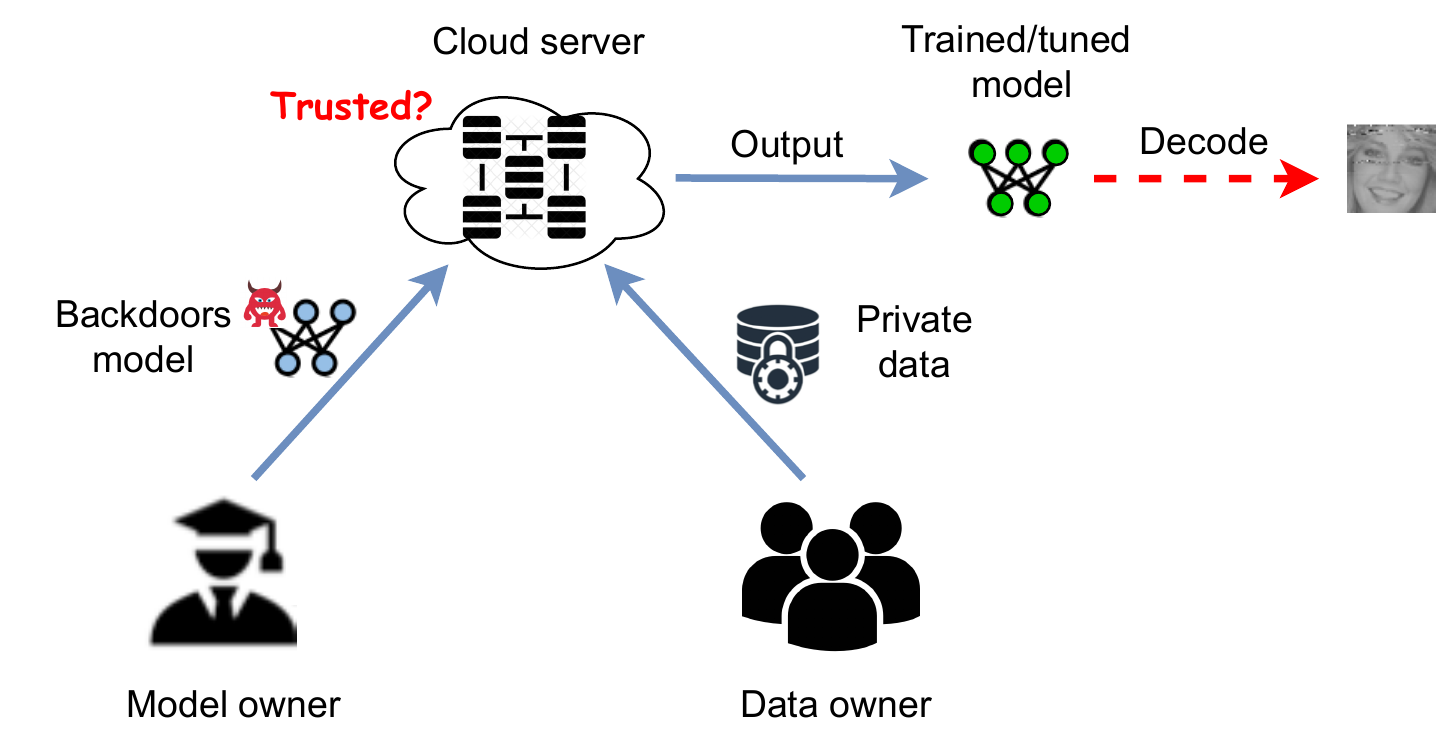}  \vskip-0.1in 
    \caption{\label{fig:third_party} \rt{Learning through third-party ML platforms suffers from data extraction attacks~\cite{song2017machine} in which model owners provide backdoors models even if the cloud server is trusted.}}
\end{figure}

\rt{\emph{Real case II.} Learning without third-party ML platforms. In ML marketplaces, a third-party cloud service with isolated environments~\cite{zhai2016cqstr} works as Fig.~\ref{fig:third_party}. One user (Data owner) supplies sensitive data, another (Model owner) supplies a valuable learning algorithm, and the cloud ensures that the algorithm cannot communicate with the outside world except by outputting a trained/tuned model. Even if such a third-party platform is trusted, the backdoors model can extract data by encoding the data into model parameters~\cite{song2017machine}. Our EXPECTED setting enables data and model communication without the participation of any third-party platforms. Since the model is updated on the model owners' side, the backdoors model cannot work under EXPECTED.}

\section{\ro{Experimental Comparison with Other Black-box Optimization Methods }}\label{appendix:BO_comparison}

\ro{\emph{Experimental setup.} CMA-ES is one of the most powerful BO methods of which the covariance matrix is updated to increase the probability of successful candidate solutions and reduce search steps. We apply a lightweight implementation\footnote{https://pypi.org/project/cmaes/} in our experiments. Bayesian optimization (abbr. Bayesian) attempts to find the maximum value of an unknown function in as few iterations as possible. We use the implementation of Bayesian global optimization with Gaussian processes\footnote{https://github.com/fmfn/BayesianOptimization} in our experiments. To simulate the model tuning setting, we assign pre-trained model parameters as the initial samples of CMA-ES and Bayesian for a fair comparison. We also create the boundaries of optimized variables in Bayesian by measuring the distance between the pre-trained model and fine-tuned (with target data) model. Other hyperparameters are used as defaults.}

\ro{\emph{Results analyses.} Experimental results on Adult and Amazon are shown as Fig.~\ref{fig:BO_comparison}. We observe that  (a) PPS and CMA-ES produce close performance curves w.r.t. query number. Bayesian with carefully-set boundaries only slightly improves the tuning performance, and its time cost is also dramatically high. (b) From Fig.~\ref{fig:BO_comparison}(a), PPS is a bit inferior to CMA-ES when the query number is $\le 500$. One possible reason is that PPS is sensitive to step size, especially at the early stage of the tuning. We prefer PPS as it still works when tuned parameters are high-dimensional, i.e., Amazon, while the other two failed to be salable in such cases.}

\section{\rt{Relation between EXPECTED, BO and RL}}
\begin{figure}[h]
    \centering
    \includegraphics[width=0.47\textwidth]{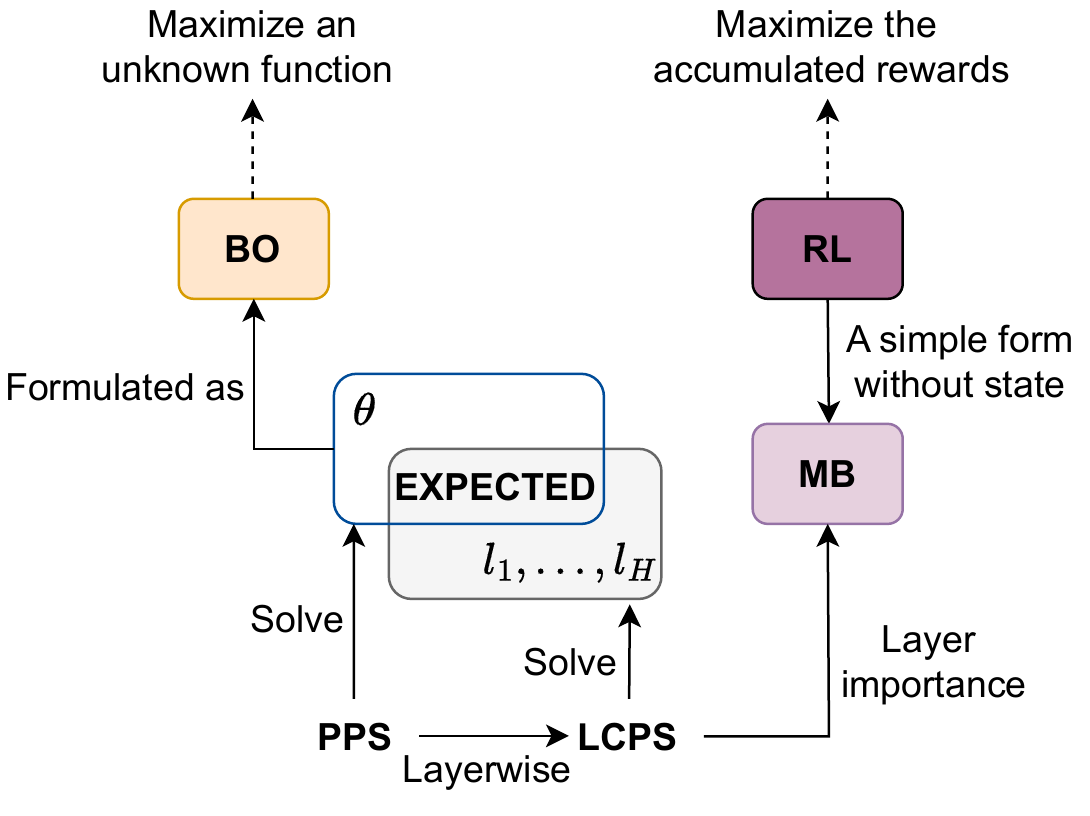}  \vskip-0.1in 
    \caption{\label{fig:relation_to_RL} \rt{The relation between EXPECTED, BO and RL.}}
\end{figure}
\rt{Fig.~\ref{fig:relation_to_RL} RL presents a general understanding of their relations. The proposed EXPECTED setting is formulated as the Black-box Optimization (BO) which ``puts target data into a black box'' instead. Performance-guided Parameter Search (PPS) is developed to address this challenge in a simple scenario where only a single variable $\theta$ is to be updated. We extend PPS to LCPS (Layerwise Coordinate Parameter Search) when multiple layers need to be updated given a fixed query budget. Since the performance gains by updating different layers are different, updating a specific layer is like choosing an arm of a bandit. That means the layer selection based on layer importance in LCPS can be understood as a Multi-armed Bandit problem (MB). B involves an exploration-exploitation trade-off and it is essentially a simple form of Reinforcement Learning (RL). They differ technically in that MB lacks an explicit model of state-to-action mapping.
In terms of their objectives, BO aims to find the solution that maximizes (or minimizes) an unknown function while RL aims to learn a policy that directs actions to obtain the accumulated rewards. Such a gap is bridged when models become complex and exploration space is expanded. From a micro view, some formulation about updating policy gradient is quite similar to PPS. One example can be found from this webpage\footnote{https://spinningup.openai.com/en/latest/spinningup/rl\_intro3.html}.}

% % you can choose not to have a title for an appendix
% % if you want by leaving the argument blank
% \section{}

% % use section* for acknowledgment
% \ifCLASSOPTIONcompsoc
%   % The Computer Society usually uses the plural form
%   \section*{Acknowledgments}
% \else
%   % regular IEEE prefers the singular form
% \fi


\ifCLASSOPTIONcaptionsoff
  \newpage
\fi

\end{document}